\PassOptionsToPackage{table}{xcolor}
\documentclass[runningheads]{llncs}

\usepackage{eccv}

\usepackage{eccvabbrv}

\usepackage{graphicx}
\usepackage{booktabs}

\usepackage{relsize}

\usepackage[accsupp]{axessibility}  

\usepackage[pagebackref,breaklinks,colorlinks,citecolor=eccvblue]{hyperref}

\usepackage{orcidlink}
\usepackage{anyfontsize}
\usepackage{ stmaryrd }
\usepackage{amsmath,amsfonts,bm}
\usepackage{cuted}
\usepackage[ruled,vlined,linesnumbered]{algorithm2e}
\usepackage{graphicx}
\usepackage{tabularx} 

\usepackage{amssymb}
\usepackage{multirow}

\usepackage{floatrow}
\usepackage{color}
\usepackage[bb=boondox]{mathalfa}
\usepackage{dsfont}
\usepackage{upgreek}
\usepackage{mathrsfs}
\usepackage{url}

\usepackage{pifont}
\def\eqref#1{equation~\ref{#1}}

\def\1{\bm{1}}

\DeclareMathAlphabet{\mathsfit}{\encodingdefault}{\sfdefault}{m}{sl}
\SetMathAlphabet{\mathsfit}{bold}{\encodingdefault}{\sfdefault}{bx}{n}

\usepackage{xcolor}
\usepackage[table]{xcolor}
\usepackage{cite}
\usepackage{latexsym}

\usepackage{flushend}
\usepackage{lipsum}

\usepackage{booktabs}
\newcommand{\improvecolor}{\color[HTML]{3B9612}}

\title{Efficient Human-Contact Representation for Human-Scene Interaction}

\author{
    Nghia Vu$^{1}$, Tuong Do$^{1,2,3}$, Binh X. Nguyen$^{1}$, Erman Tjiputra$^{1}$, Anh Nguyen$^{2}$%
}

\authorrunning{N.~Vu et al.}

\institute{AIOZ, Singapore 
\and
Department of Computer Science, University of Liverpool, UK
\and 
Department of Computer Science, NTHU, Taiwan
}

\begin{document}
\maketitle

\begin{table*}
\vspace{-0.8 cm}
\resizebox{\linewidth}{!}{
\setlength{\tabcolsep}{2pt}
\begin{tabular}{cccc}
\shortstack{\includegraphics[width=0.24\linewidth,height=0.21\linewidth]{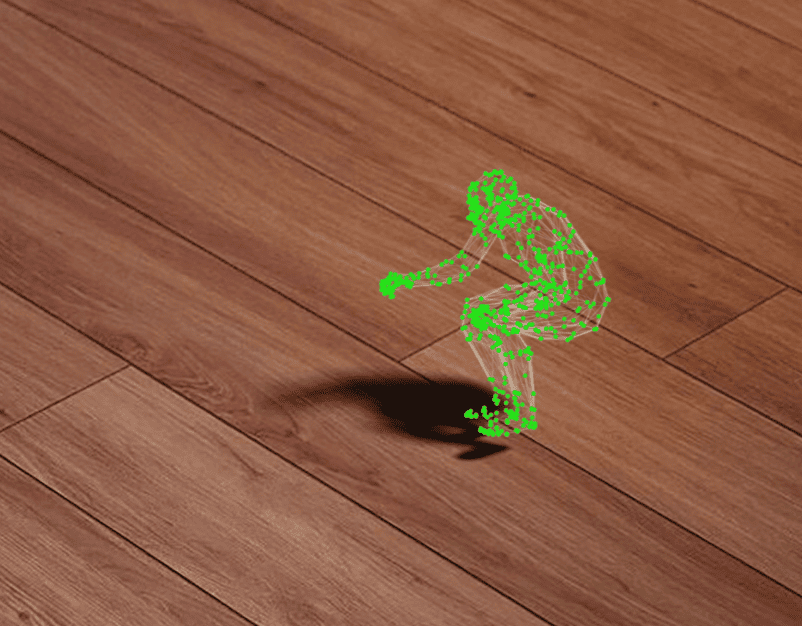}}&
\shortstack{\includegraphics[width=0.24\linewidth,height=0.21\linewidth]{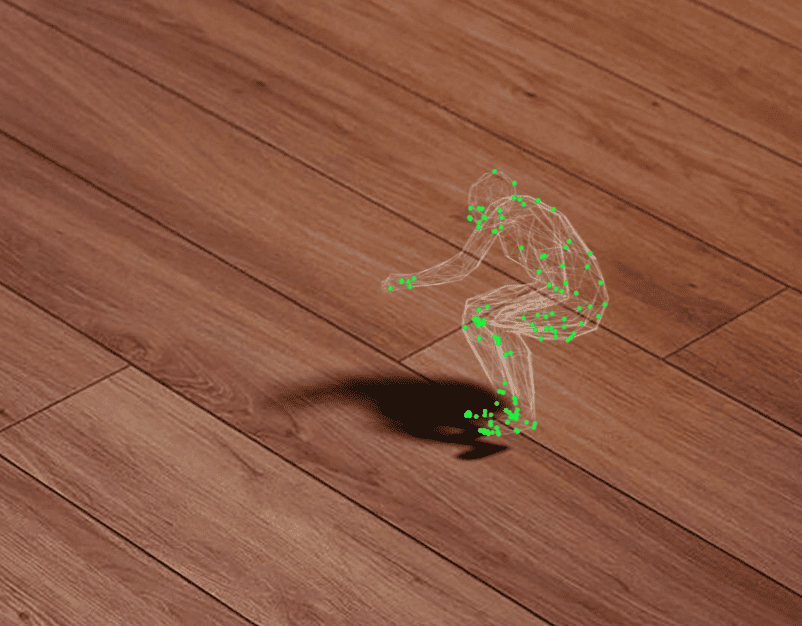}}&
\shortstack{\includegraphics[width=0.24\linewidth,height=0.21\linewidth]{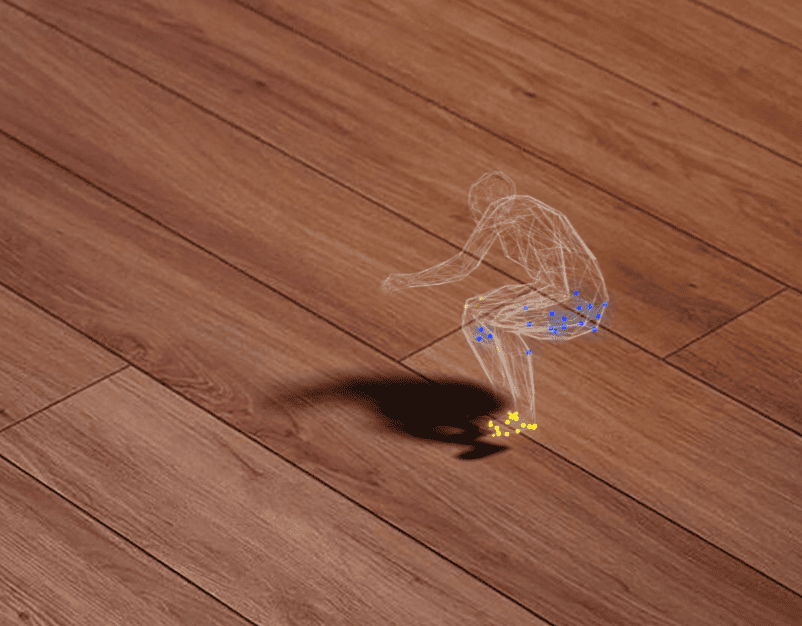}}&
\shortstack{\includegraphics[width=0.24\linewidth,height=0.21\linewidth]{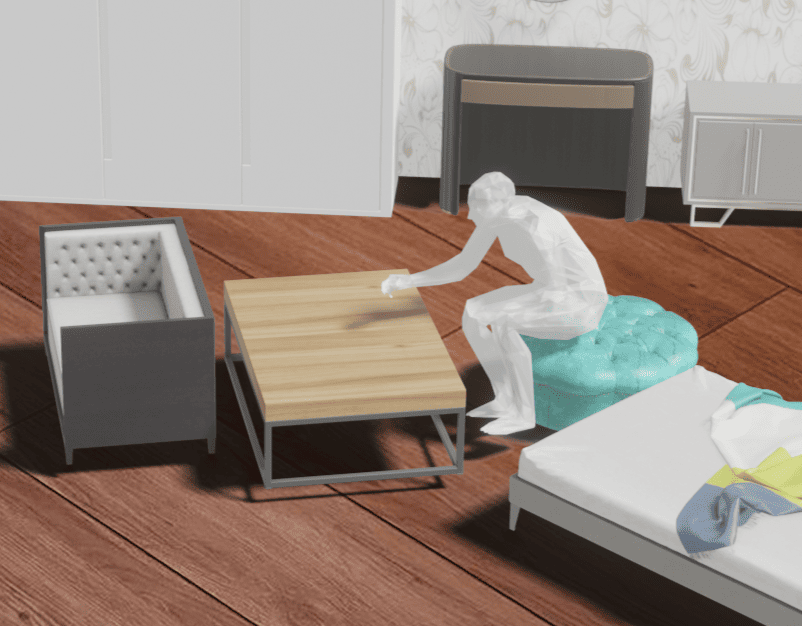}}\\[1pt]
\shortstack{\small (a) Input}&
\shortstack{\small (b) Human-Contact}&
\shortstack{\begin{tabular}[c]{@{}c@{}}\small  (c) Contact Predict\end{tabular}}&
\shortstack{\begin{tabular}[c]{@{}c@{}}\small (d) Scene Synthesis\end{tabular}}
\end{tabular}
}
 \captionof{figure}{\textbf{Efficient contact representation for human-scene interaction}. Given dense input (a), we learn and select useful information to create efficient contact representation (b), for contact prediction (c) and scene synthesis tasks (d).}
\label{fig:teaser}
\vspace{-0.1cm}
\end{table*}
\vspace{-1.2 cm}

\begin{abstract}
Human-scene interaction is an active research topic with several industrial applications in virtual reality, gaming, robotics, and surveillance. Despite significant progress in network architectures to improve the results or optimize models' parameters for fast inference speed, the efficient representation of contact between humans and their environments remains an open challenge. In this paper, we propose a new efficient human-contact representation for human-scene interaction. Our primary contribution is the introduction of sparse contact masks that strategically select essential contact information, significantly reducing redundant data in high-dimensional inputs. Leveraging this efficient contact representation, we propose a suite of sparse operators to replace traditional dense operators within deep network layers for faster computation. Our approach not only enhances computational speed but also filters out non-essential contact data, thereby improving the precision of human-scene interaction models. To validate the effectiveness of our method, we conduct intensive experiments across three public benchmark datasets, focusing on two critical tasks for human-scene interaction: contact prediction and scene synthesis. The experimental results show that our approach outperforms state-of-the-art models in reconstruction accuracy and achieves a computation speed-up of at least 12 times over recent baselines.
\end{abstract}
\section{Introduction}



Human-scene interaction explores how humans perceive, navigate, and engage with the environment around them~\cite{hassan2021populatingPOSA,tanke2024humans, petrov2023object}. Recently, there has been significant attention on learning the dynamics between humans and the environment~\cite{zhang2022couch,yi2023mime, cheng2024towards}. To enhance the modeling and understanding of human pose within diverse environments, researchers have investigated several topics such as human-scene interactions~\cite{hassan2021populatingPOSA,luo2023leverage, mullen2023placing, ye2023affordance, jiang2024scaling}, human-scene synthesis~\cite{zhao2022compositional,shen2023learning,blinn2021learning,vuong2023language, tripathi20233d}, or human pose contact prediction~\cite{zheng2022gimo,huang2022capturingRICH}. Gaining a comprehensive understanding of human posture and interactions with the environment is crucial for various downstream applications~\cite{ye2022scene} such as human-robot interaction~\cite{MANO:SIGGRAPHASIA:2017,yi2022human}, realistic virtual experiences~\cite{arsalan2017synthesizing,zhao2022compositional, qiu2023virtualhome}, game animations~\cite{habermann2021real}, intuitive interfaces~\cite{zou2018sketchyscene}, advanced surveillance systems~\cite{benfold2009guiding}, and healthcare applications~\cite{meng2023virtual}.

In human-scene interaction, numerous approaches concentrate on generating high-quality scenes based on human contacts and interactions~\cite{hassan2021populatingPOSA,wang2022reconstructing,jiang2022chairs,zheng2022gimo,yi2022human,ye2022scene,wang2022towards,yi2023mime}. While the development of complex networks capable of handling the intricacies of scene generation tasks is essential, it also poses challenges in terms of inference speed~\cite{lee2023multi} and effectively process the data~\cite{NEURIPS2022_a0303731}. 
Yet, many works have acknowledged this problem and focused on lightweight architectures, model pruning, or quantization to improve model accuracy and enhance inference speed~\cite{riegler2017octnet,zhang2022couch,schwarz2020stillleben}.
However, despite developments, current methods still struggle to process complex contact interactions between the temporal-spatial dynamics of human poses and surrounding objects in scenes. 

In this paper, unlike previous methods that primarily focus on designing lightweight models, quantization, model pruning, or diffusion models to enhance human-scene interaction~\cite{hassan2019resolvingPROX,liu2022efficient, Jiang_2022_CVPR}, we propose a solution that focuses on effectively representing the \textit{contact between humans and the scene}. We are motivated by the fact that the input data for human-scene interaction are complex but have sparse structures, while having an effective way to represent the contact has shown significant improvement in terms of both accuracy and inference speed in other tasks such as affordance learning~\cite{morais2021learning, Bahl_2023_CVPR} or NeRF-based scene generation~\cite{zhao2022humannerf,niemeyer2022regnerf, zhang2023nerflets}. In particular, we propose Efficient Contact Representation (ECO) for human-scene interaction. Our method utilizes a set of sparse contact masks to effectively select important information from the human-scene data (Fig.~\ref{fig:teaser}). We then propose a set of sparse operators to replace traditional dense tensors in deep network layers. Intensive experiments show that our method outperforms recent works in contact prediction and scene synthesis tasks while achieving much faster inference speed.
\section{Related Work}

\textbf{Human-Scene Contact Modeling.}
Human-scene interaction aims to understand how humans physically engage with surrounding environments and how these interactions can be represented, predicted, and synthesized~\cite{li2019putting}. Advances in parametric human body models, including SMPL~\cite{loper2015smpl}, SMPL-X~\cite{SMPL-X:2019}, MANO~\cite{MANO:SIGGRAPHASIA:2017}, and FLAME~\cite{FLAME:SiggraphAsia2017}, have enabled a wide range of methods for modeling contact-rich interactions between humans and scenes. Early works focused on learning affordances and plausible human poses from visual observations~\cite{wang2017binge,li2019putting}. To support large-scale learning, several datasets have been introduced, including VirtualHome~\cite{puig2018virtualhome}, which provides simulated environments for embodied agents, and BEHAVE~\cite{bhatnagar2022behave}, which captures real-world human-object interactions with detailed contact annotations.

Building upon these datasets, recent methods have addressed scene population~\cite{hassan2021populatingPOSA,wang2022reconstructing,jiang2022chairs}, affordance learning~\cite{kulal2023puttingAffordance1,luo2023leverage,wang2024move,zhang2025iaao}, human-object and full-body interaction modeling~\cite{MANO:SIGGRAPHASIA:2017,SMPL-X:2019,meng2025rethinking,cong2025semgeomo}, human-scene generation~\cite{nie2022pose2room,wang2022humanise,li2024genzi,cen2024generating,hwang2025scenemi,huang2025move,guo2025motionlab}, diffusion-based scene synthesis~\cite{zheng2022gimo,yi2022human,ye2022scene,wang2022towards,yi2023mime,wei2025functional,yang2026sceneweaver}, and interaction tracking~\cite{blinn2021learning,yi2022physicalPIP,xie2023visibility,xu2025interdreamer,lu2025humoto,hong2026learning,xu2026interprior}. These advances contribute significantly to human-object interaction understanding, scene generation, and embodied reasoning~\cite{zhang2020perceiving,wang2022towards,jiang2024autonomous,liu2025tcpformer,zhou2025hermes,deng2026gaussiandwm,huang2026surprise3d}. Despite their success, most existing methods rely on dense contact representations that process all body vertices equally, even though only a small subset of vertices actively participates in physical interactions. Consequently, the redundancy inherent in dense human-scene representations remains largely unexplored.

\textbf{Efficient Human-Scene Interaction Models.}
Improving the efficiency of human-scene interaction models has attracted increasing attention due to the computational complexity of processing high-dimensional human and scene representations. Existing approaches primarily focus on reducing network complexity through model pruning~\cite{netpruning}, redundancy reduction~\cite{exploit_redundancy}, quantization~\cite{liu2018bireal,liu2019bireal}, knowledge distillation~\cite{hinton2015distilling}, and neural architecture search~\cite{nas1,nas2,nas3,nas4}. Similar strategies have been explored in related applications such as trajectory prediction~\cite{liu2022efficient,katariya2022deeptrack} and dynamic scene generation~\cite{su2022robustfusion,arad2021compositional,wang2025terrain,chu2024dreamscene4d}. While these methods effectively reduce computational cost, they primarily optimize the network architecture while leaving the underlying human-scene representation unchanged. As a result, redundant input information is still propagated throughout the network. In contrast, our work improves efficiency at the representation level by identifying and preserving only the most informative contact regions before network processing, thereby reducing both computational overhead and representation redundancy.

\textbf{Sparse Representations for Geometric Data.}
Beyond architectural optimization, another direction for accelerating inference is to exploit sparsity within the input representation itself. Prior works have investigated sparse tensor representations~\cite{choy20194dMinkowskiME,choy2020high}, sparse convolutional operators~\cite{graham20173d,escoin}, sparse matrix acceleration~\cite{sylos2022blocking}, mesh simplification~\cite{alliez2003isotropic,rossignac1993multi,ghazanfarpour2020proximity,potamias2022neural,ranjan2018generating}, and compact geometric representations~\cite{edavamadathil2024neural}. Graham et al.~\cite{graham20173d} introduce submanifold sparse convolutions to efficiently process spatially sparse data, while Chen et al.~\cite{escoin} design sparse-kernel convolutions with optimized memory access patterns. Labini et al.~\cite{sylos2022blocking} propose blocking strategies for sparse matrix multiplication with theoretical guarantees on density preservation. More recently, Edavamadathil et al.~\cite{edavamadathil2024neural} introduce Neural Geometry Fields that represent meshes using a compact set of quadrangular patches.

Although sparse representations have demonstrated effectiveness across geometric learning tasks, their application to human-scene interaction remains limited. Existing sparse methods primarily exploit geometric sparsity, whereas interaction reasoning depends on semantically meaningful contact regions that govern physical human-scene interactions. Directly applying generic sparsification may therefore discard important interaction cues. To address this challenge, we introduce a contact-aware sparse representation that explicitly identifies informative contact regions through sparse contact masks and integrates them with sparse operators. By exploiting semantic contact sparsity rather than purely geometric sparsity, our approach reduces redundancy while preserving interaction-critical information, leading to both improved efficiency and stronger human-scene interaction modeling.

\begin{figure*}[!ht]
\vspace{-0.4 cm}
   \centering
   \includegraphics[width=1\linewidth]{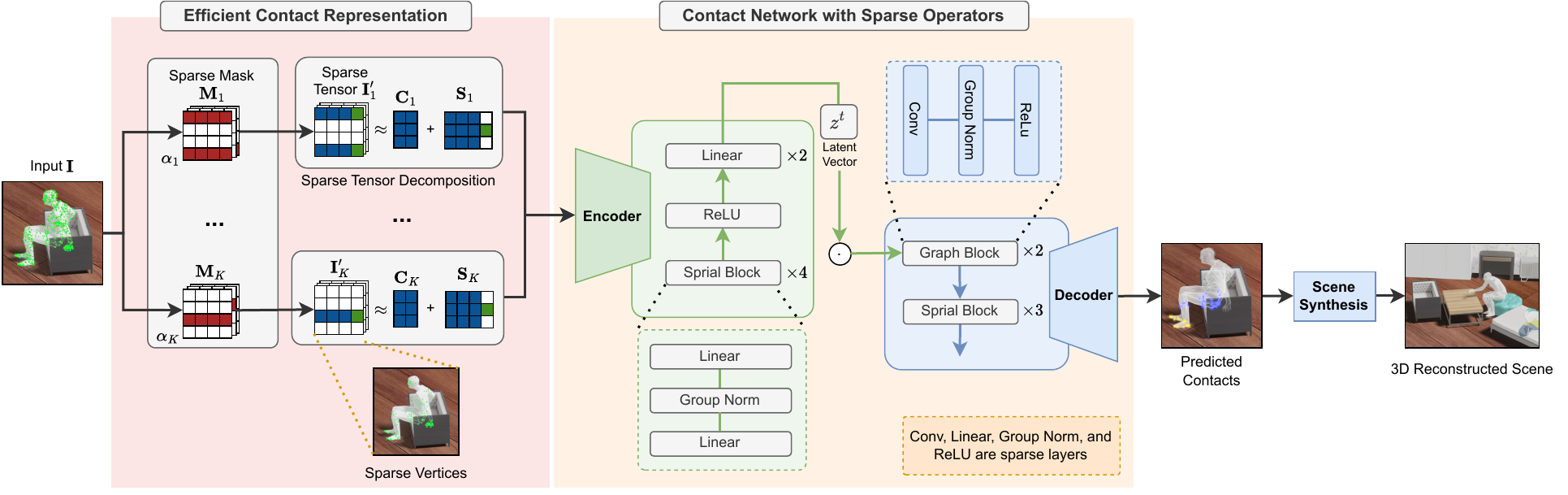}
 \caption{\textbf{Efficient Contact Representation method overview}. 
 We adopt Graph and Spiral Blocks from POSA~\cite{hassan2021populatingPOSA} as contact predictor backbone, replacing original layers with our sparse layers. 
 Red cells denote non-zero kernel weights and mask values, blue cells represent coordinates, green cells indicate non-zero contact values, and white cells denote zeros.
 \vspace{-0.3 cm}
 }
 \label{fig:SMSmethod}
\vspace{-0.3 cm}
\end{figure*}
\vspace{-0.8 cm}
\section{Efficent Contact Representation}

\subsection{Contact Representation with Sparse Mask}
\label{subsec:ECO}

We follow~\cite{hassan2021populatingPOSA} to represent the human-scene contact. In particular, the human-scene input tensor $\bm I$ is defined as $\bm I =(\bm V, \bm F)$, where $\bm V \in \mathds{R}^{N_v \times 3}$ is body vertices and $\bm F\in \mathds{R}^{N_v \times N_c}$ is the contact label of the vertices. $N_v$ is the number of vertices, and $N_c$ is the number of labels.

\textbf{Sparse Mask.} Our goal is to convert the human-scene input tensor $\bm{I}$ into a sparse tensor $\bm{I'}  \in \mathds{R}^{N_v \times N_S}$ for a more efficient contact representation ($N_S = N_c + 3$). We define a sparse mask $\bm{M} \in \mathds{R}^{N_v \times N_S}$ and calculate $\bm{I'}$ as:
\begin{equation}
    \bm{I'} = \bm{M} \circ \bm{I}
    \label{eq:mask_main}
\end{equation}
where $\circ$ denotes element-wise multiplication. Each element in the sparse mask $\bm{M}$ is sampled from a binomial distribution. The sparsity of $\bm{M}$ is controlled via a \textit{sparsity ratio} parameter which indicates the non-zero value ratio of the mask. Intuitively, $\bm{M}$ is a matrix with only $0$ or $1$ values to mask out the unnecessary information from the input.





In practice, applying only a \textit{single} high-sparsity mask $\bm{M}$ to the input causes significant information loss hence heavily affecting the effectiveness of the model. To overcome this limitation, we apply $K$ \textit{multiple} sparse masks $\left\{\bm{M}_1, \bm{M}_2, ..., \bm{M}_K \right\} $ to the input with the expectation that each sparse mask $\bm{M}_k$ would learn different important information from the input. We note that each sparse mask $\bm{M}_k$ is applied independently to the input to obtain the sparse tensor $\bm{I'}_k$, and $K$ is the hyper-parameter that indicates how many sparse masks we use during training.


After applying the sparse mask $\bm{M}_k$ to the input tensor $\bm{I}$, we obtain a sparse tensor $\bm{I'}_k = \bm{M}_k \circ \bm{I}$ which has a high proportion of zero values. Consequently, the conventional dense representation is inefficient for representing the sparse tensor $\bm{I'}_k$ during the learning process. Additionally, effectively storing only non-zero values in the sparse tensor facilitates computation~\cite{tew2016investigation}. To this end, we \textit{decompose} the sparse tensor $\bm{I'}_k$ into two tensors to remove the zero values as in~\cite{chou2018format}. This decomposition results in a coordinate matrix $\bm{C'}_k \in \mathds{R}^{N'_k \times 2}$ and an associated feature matrix $\bm{S'}_k \in \mathds{R}^{N'_k \times N'_S}$  where $N'_k$ denotes the number of non-zero values in $\bm{I'}_k$. This strategy not only saves memory by removing zero-values from the sparse tensor but also streamlines the computation process for $\bm{I'}_k$. In practice, the sparse tensor $\bm{I'}_k $ is represented as $\bm{I'}_k = \left (  \bm{C'}_k | \bm{S'}_k \right)$, where $\bm{C'}_k$ and $\bm{S'}_k$ are defined as: 
 \begin{equation}
 \bm{C'}_k = \begin{bmatrix}
 b_1  & x_1 \\ 
 \vdots &  \vdots\\ 
  b_{N'_k}  & x_{N'_k}  
\end{bmatrix} , \bm{S'}_k = \begin{bmatrix}
\bm{s}_1^{\intercal}\\ 
\vdots\\
\bm{s}_{N'_k}^{\intercal}
\end{bmatrix} 
\end{equation}
where $\left (b_i, x_i\right )$ is the frame index and coordinate of $i$-th feature $\bm{s}_i \in \mathds{R}^{N'_S}$. 



\textbf{Sparse Mask Selection.} Although using a list of sparse masks preserves the model's performance compared to using a single mask, it leads to the fact that some sparse masks capture duplicate information or unnecessary features in the input which may have a negative effect on the results or slow down the inference. To resolve this problem, we define the learnable \textit{mask score} $\bm{\alpha} \in \mathds{R}^K$ to \textit{indicate the importance} of each sparse mask. This mask score is calculated based on the contribution of each mask to the final results and the similarity between corresponding masks as follows: 
\vspace{-1ex}
\begin{equation}
\bm{\alpha}_{(t+1,k)}
= \bm{\alpha}_{(t,k)} + \frac{1}{K-1}\sum_{i \neq k, 1 \leq i \leq K} \left(1-  \frac{\left \| {\bm{O}^{\intercal}_{(t,i)}}\bm{O}_{(t,k)} \right \|_\text{F}^2}{\left \| \bm{O}_{(t,k)}^{\intercal}\bm{O}_{(t,k)} \right \|_\text{F} \left \| \bm{O}_{(t,i)}^{\intercal}\bm{O}_{(t,i)} \right \|_\text{F}}\right)
\label{eq:ECO}
\end{equation}
where $\left \| . \right \|_{\text{F}}$ is the Frobenius norm; $t$ corresponds to iteration during learning; $\bm{O}_k$ is the output tensor corresponding to mask $\bm{M}_k$. Our goal is to compare the differences in distribution between features outputted from different sparse masks to identify which masks mostly produce the same outputs and then discard the redundant ones during the inference process. We note that during training, we utilize $K$ sparse masks and calculate the associated mask scores, while \textit{during testing, we select $\kappa$ masks} ($\kappa << K$) based on the mask score $\bm \alpha$ to use only the useful masks. The selected useful masks are then applied to the input human-scene representation (Equation~\ref{eq:mask_main}) to produce the efficient contact presentation for human-scene interaction.

\subsection{Contact Network with Sparse Operations}
\label{app_sec:sparselayer}
Traditional contact networks such as POSA~\cite{hassan2021populatingPOSA} learn the human-scene contact using the full dense input tensor $\bm{I}$ with conventional layers such as convolution, group normalization, ReLU, etc. This leads to two problems: \textit{i)} the dense input $\bm{I}$ contains unnecessary information (e.g., non-contact points) which may decrease the accuracy of the network, and \textit{ii)} learning on a dense input $\bm{I}$ reduces the inference time as whole tensor $\bm{I}$ is used and subsequently increases the number of parameters of the network. To utilize our efficient contact representation $\bm{I'}$, we propose to replace the conventional matrix operations with our designed sparse operations, utilizing input from our sparse mask. This strategy can be applied across different layers, including convolution, batch normalization, pooling, and more, all without necessitating changes to the network architecture. Next, we describe sparse operations in popular deep network layers.

\subsubsection{Convolution Layer}
The $k$-th reformatted inputs $\bm{S}_k$ and $\bm{C}_k$ are passed through the network and interact with sparse kernels $\bm{W} \in  \mathds{R}^{m \times m}$ via a mapping function. In the convolutional layer, kernel weights with an indexing matrix $\mathcal{M}^n_k$ of a $k$-th mask at the $n$-th stride can be calculated as follows:
\begin{equation}
\mathcal{M}^n_k = \begin{bmatrix} \hat{\bm{W}}[c] \ | \  \bm{C}_k[c'] \\\hat{\bm{W}}[i] \ | \ \bm{C}_k[i'] \end{bmatrix}, \ \ \begin{matrix}
c = (m^2-1)/2, \  c' = i' \  \forall i = c \\ i' = \text{idx}(\bm{C}_k[\bm{W},n,i]), \  \hat{\bm{W}}[i] \neq 0
\end{matrix}
\label{eq:mapping}
\end{equation} 
where $\hat{\bm{W}}$ is the flattened vector of the kernel $\bm{W}$ and $\bm{C}_k[\bm{W},n,i]$ is the value when kernel $\bm{W}$ is applied to the $k$-th sparse input $\bm{C}_k$ over $n$-th stride corresponding to the $i$ element. $\mathcal{M}^n_k$ is then retrieved in $\bm{C}_k$ and $\bm{S}_k$ to compute the sparse output  $\bm{C'}_k$ and  $\bm{S'}_k$ using Equation~\ref{eq:retrieve}.
\begin{equation}\label{eq:retrieve}
\begin{gathered}
\bm{C}^{'n}_k =  \left [ \mathcal{M}_k^n[0][1:] \right], \\ 
\bm{S}^{'n}_k = \sum_{i=1} \mathcal{M}_k^n[i][0] \bm{S}_k[\text{idx}\left ( \mathcal{M}_k^n[i][1] \right)]
\end{gathered}
\end{equation}

\subsubsection{Linear Layer}
Linear layers applied to sparse tensors only change the number of channels in the feature matrix and do not affect the coordinate matrix. With $\bm{W}_l \in \mathds{R}^{N_S \times N_S'}$ and $\bm{b} \in \mathds{R}^{N_S'}$ are the weight matrix and bias vector of the linear layer, respectively, the linear operator output is:
\begin{equation}
    \bm{S'}_{k} = \bm{W}_l\bm{S}_{k} + \bm{b}, \ \ \bm{C'}_k = \bm{C}_k 
\end{equation}

\subsubsection{Group Normalization Layer}
In group normalization, we divide the features into $G$ group, each group has $N_v / G$ feature values. Then the values in each group are normalized: 
\begin{equation}
    \bm{\mu}_{k,g}^b = \dfrac{1}{N_k^b} \sum_{\substack{i:\bm{C}_k[i][0] = b}} \bm{S}_{k,g}[i]
\end{equation}
\begin{equation}
    (\bm{\sigma}_{k,g}^b)^2 =  \dfrac{1}{N_k^b}\sum_{i:\bm{C}_k[i][0] = b} \left (  \bm{S}_{k,g}[i] - \bm{\mu}_{k,g}^b \right )^2 
\end{equation}
\begin{equation}
    \bm{S}^{'b}_{k,g}
    = \dfrac{\bm{S}^b_{k,g} - \bm{\mu}^b_{k,g}}{\sqrt{(\bm{\sigma}^b_{k,g})^2 + \epsilon}}, \ \ \bm{C'}_k = \bm{C}_k
\end{equation}
where $g$ is the group index. When $G=1$, the group normalization becomes layer normalization instead.

\subsubsection{ReLU Layer}
For the ReLU and any other non-linearly layers, sparse operations only change each value in the feature matrix and do not affect the coordinate matrix. 
With  the activation function $\bm{f_{\text{act}}}$, the output is calculated as:
\begin{equation}
    \bm{S'}_k = \bm{f_{\text{act}}}\left( \bm{S}_k \right) , \ \ \bm{C'}_k = \bm{C}_k 
\end{equation}

\begin{table}[ht]
\vspace{-0.7 cm}
\centering
\resizebox{\linewidth}{!}{
\begin{tabular}{r|cccccc|c}
\hline
\multirow{4}{*}{\textbf{Methods}} & \multicolumn{6}{c|}{\textbf{Datasets}} & \multirow{4}{*}{\textbf{\begin{tabular}[c]{@{}c@{}}Inference\\ Speed\\ (s/sample)\end{tabular}}} \\ \cline{2-7}
 & \multicolumn{2}{c|}{\textit{\textbf{PROXD}}} & \multicolumn{2}{c|}{\textit{\textbf{GIMO}}} & \multicolumn{2}{c|}{\textit{\textbf{BEHAVE}}} &  \\ \cline{2-7}
 & \multicolumn{1}{c|}{\textit{\begin{tabular}[c]{@{}c@{}}Reconstruction\\ Accuracy (\%)\end{tabular}}} & \multicolumn{1}{c|}{\textit{\begin{tabular}[c]{@{}c@{}}Consistency\\ Score\end{tabular}}} & \multicolumn{1}{c|}{\textit{\begin{tabular}[c]{@{}c@{}}Reconstruction\\ Accuracy (\%)\end{tabular}}} & \multicolumn{1}{c|}{\textit{\begin{tabular}[c]{@{}c@{}}Consistency\\ Score\end{tabular}}} & \multicolumn{1}{c|}{\textit{\begin{tabular}[c]{@{}c@{}}Reconstruction\\ Accuracy (\%)\end{tabular}}} & \textit{\begin{tabular}[c]{@{}c@{}}Consistency\\ Score\end{tabular}} &  \\ \hline
LSTM
& \multicolumn{1}{c|}{\begin{tabular}[c]{@{}c@{}}90.91 \improvecolor{(+2.78)}\end{tabular}} & \multicolumn{1}{c|}{\begin{tabular}[c]{@{}c@{}}0.921 \improvecolor{(+0.06)}\end{tabular}} & \multicolumn{1}{c|}{83.2 \improvecolor{(+8.9)}} & \multicolumn{1}{c|}{0.814 \improvecolor{(+0.129)}} & \multicolumn{1}{c|}{80.8 \improvecolor{(+13.0)}} &0.766 \improvecolor{(+0.107)}  & 0.17
\\ 
POSA
& \multicolumn{1}{c|}{91.12 \improvecolor{(+2.57)}} & \multicolumn{1}{c|}{0.882 \improvecolor{(+0.099)}} & \multicolumn{1}{c|}{89.9 \improvecolor{(+2.2)}} & \multicolumn{1}{c|}{0.909 \improvecolor{(+0.034)}} & \multicolumn{1}{c|}{89.7 \improvecolor{(+4.1)}} &0.854 \improvecolor{(+0.019)}  & 0.28
\\
ContactFormer
& \multicolumn{1}{c|}{91.27 \improvecolor{(+2.42)}} & \multicolumn{1}{c|}{0.952 \improvecolor{(+0.029)}} & \multicolumn{1}{c|}{90.7 \improvecolor{(+1.4)}} & \multicolumn{1}{c|}{0.912 \improvecolor{(+0.031)}} & \multicolumn{1}{c|}{91.1 \improvecolor{(+2.7)}} &0.845 \improvecolor{(+0.028)}  & 0.20 
\\ 
MIME
& \multicolumn{1}{c|}{90.97 \improvecolor{(+2.72)}} & \multicolumn{1}{c|}{0.902 \improvecolor{(+0.079)}} & \multicolumn{1}{c|}{89.9 \improvecolor{(+2.2)}} & \multicolumn{1}{c|}{0.911 \improvecolor{(+0.032)}} & \multicolumn{1}{c|}{90.2 \improvecolor{(+3.6)}} & 0.854 \improvecolor{(+0.019)} & 0.54
\\
PIAL-Net
& \multicolumn{1}{c|}{92.04 \improvecolor{(+1.65)}} & \multicolumn{1}{c|}{0.953 \improvecolor{(+0.028)}} & \multicolumn{1}{c|}{91.1 \improvecolor{(+1.0)}} & \multicolumn{1}{c|}{0.934 \improvecolor{(+0.009)}} & \multicolumn{1}{c|}{89.9 \improvecolor{(+3.9)}} &0.864 \improvecolor{(+0.009)}  & 2.97
\\
HOT
& \multicolumn{1}{c|}{90.9 \improvecolor{(+2.79)}} & \multicolumn{1}{c|}{0.966 \improvecolor{(+0.015)}} & \multicolumn{1}{c|}{90.3 \improvecolor{(+1.8)}} & \multicolumn{1}{c|}{0.900 \improvecolor{(+0.043)}} & \multicolumn{1}{c|}{91.7 \improvecolor{(+2.1)}} &0.821 \improvecolor{(+0.052)}  & 1.12
\\ 
TRUMANS
& \multicolumn{1}{c|}{90.7 \improvecolor{(+2.99)}} & \multicolumn{1}{c|}{0.962 \improvecolor{(+0.019)}} & \multicolumn{1}{c|}{90.9 \improvecolor{(+1.2)}} & \multicolumn{1}{c|}{0.889 \improvecolor{(+0.054)}} & \multicolumn{1}{c|}{90.4 \improvecolor{(+3.4)}} &0.844 \improvecolor{(+0.029)}  & 2.47 
\\ \hline\hline
\rowcolor[HTML]{EFEFEF}\textbf{ECO (ours)} & \multicolumn{1}{c|}{\textbf{93.69}} & \multicolumn{1}{c|}{\textbf{0.981}} & \multicolumn{1}{c|}{\textbf{92.1}} & \multicolumn{1}{c|}{\textbf{0.943}} & \multicolumn{1}{c|}{\textbf{93.8}} &\textbf{0.873}  & \textbf{0.009 }\\ \hline
\end{tabular}
}
\caption{\textbf{Quantitative comparison of ECO against different contact prediction methods}. on the PROXD, GIMO, and BEHAVE datasets. We report reconstruction accuracy, consistency score, and average inference speed per sample on the PROXD, GIMO, and BEHAVE datasets. Across all datasets, ECO consistently achieves the best reconstruction accuracy and consistency score while requiring substantially lower inference time. 
\label{tab:sota}
\vspace{-0.7 cm}
}
\vspace{-0.2 cm}
\end{table}
\vspace{-0.3 cm}
\section{Experiments}
We validate our method on two tasks: contact prediction and scene synthesis. For contact prediction, we train a conditional Variational Autoencoder (cVAE) model as in POSA~\cite{hassan2021populatingPOSA}, replacing each traditional layer with our sparse layers. For scene synthesis, we use the predicted contact labels from our efficient contact model and follow~\cite{ye2022scene} to generate objects that make contact with the human body at predicted points, ensuring alignment with human intent and avoiding body penetration.

\begin{table*}[t]
\RawFloats
\centering
\small

\begin{minipage}[t]{0.5\textwidth}
\centering

\includegraphics[width=\linewidth]{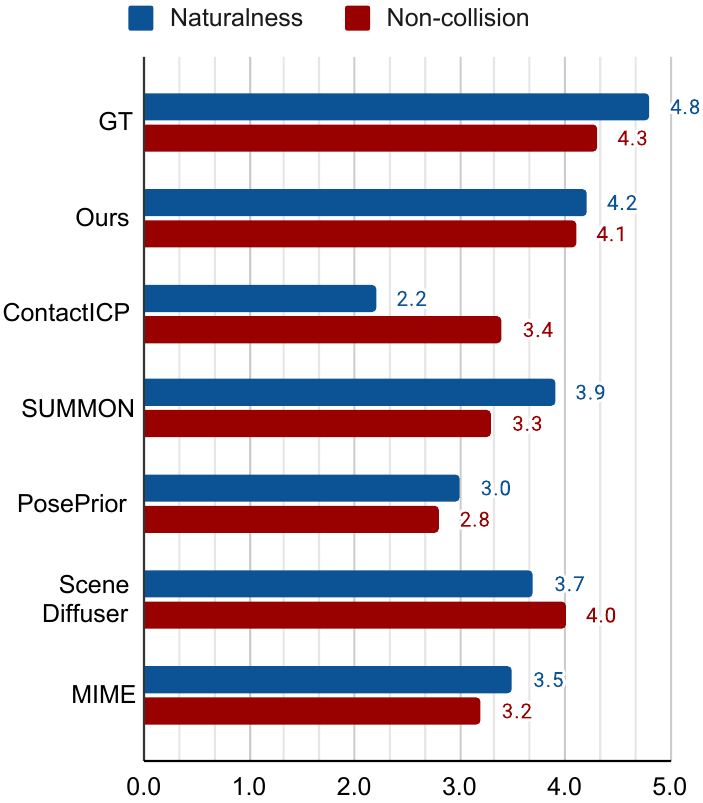}

\captionof{figure}{
User evaluation between methods.
}
\vspace{-0.2 cm}
\label{fig:UserStudy}

\end{minipage}
\hfill
\begin{minipage}[t]{0.48\textwidth}
\centering
\vspace{-7cm}
\setlength{\tabcolsep}{0.25em}
\renewcommand{\arraystretch}{1.05}

\resizebox{\linewidth}{!}{
\begin{tabular}{r|c|c}
\hline
\multirow{2}{*}{\textbf{Methods}} & \multicolumn{2}{c}{\textbf{Dataset}} \\ \cline{2-3}
&\begin{tabular}[c]{@{}c@{}}\textit{\textbf{PROXD}}\end{tabular} &
  \begin{tabular}[c]{@{}c@{}}\textit{\textbf{GIMO}}\end{tabular} \\ \hline
ContactICP
& 0.654 \improvecolor{(+0.332)}  & 0.820 \improvecolor{(+0.149)}    \\ 
PosePriors
& 0.703 \improvecolor{(+0.283)}  & 0.798 \improvecolor{(+0.171)}    \\
SUMMON
& 0.851 \improvecolor{(+0.135)}  &0.951 \improvecolor{(+0.018)}     \\ 
MIME
& 0.897 \improvecolor{(+0.089)}  &0.938 \improvecolor{(+0.031)}    \\
SceneDiffuser
& 0.914 \improvecolor{(+0.072)}  &0.942 \improvecolor{(+0.027)}    \\
INFERACT
& 0.945 \improvecolor{(+0.041)}  &0.937 \improvecolor{(+0.032)}    \\
\hline\hline
\rowcolor[HTML]{EFEFEF} \textbf{ECO (ours)} & \textbf{0.986} & \textbf{0.969}    \\ \hline
\end{tabular}
}

\vspace{-0.2 cm}
\captionof{table}{
Scene synthesis results.
\vspace{0.6 cm}
}
\label{tab:SceneSynCompare}


\setlength{\tabcolsep}{0.15em}
\renewcommand{\arraystretch}{1.2}

\resizebox{\linewidth}{!}{
\begin{tabular}{r|c|c|c}
\hline
\textbf{Methods} &
  \begin{tabular}[c]{@{}c@{}}\textit{Reconstruction}\\ \textit{Accuracy (\%)} \end{tabular} &
  \begin{tabular}[c]{@{}c@{}}\textit{Consistency}\\ \textit{Score} \end{tabular} &
  \begin{tabular}[c]{@{}c@{}}\textit{Inference Speed} \\ \textit{(s/sample)}\end{tabular} \\
  \hline
POSA
& 91.12 \improvecolor{(+2.57)} & 0.882 \improvecolor{(+0.099)} & 0.280 \improvecolor{($\downarrow$$\times$ 31.1)}   \\ 
ME
& 83.61 \improvecolor{(+10.08)}  & 0.797 \improvecolor{(+0.184)}& \textbf{0.008 } \color[HTML]{6200C9}{($\uparrow$$\times$ 1.13)} \\

EsCoin
& 69.78 
 \improvecolor{(+23.91)} & 0.721 
 \improvecolor{(+0.260)} & 0.170  \improvecolor{($\downarrow$$\times$ 1.89)}  \\

pSConv
& 90.24 \improvecolor{(+3.45)} & 0.825 \improvecolor{(+0.156)} & 0.084 \improvecolor{($\downarrow$$\times$ 9.33)}   \\

1-D Blocking
& 88.77 \improvecolor{(+4.92)} & 0.912 \improvecolor{(+0.069)} & 0.150 \improvecolor{($\downarrow$$\times$ 16.7)}    \\ \hline \hline
\rowcolor[HTML]{EFEFEF}\textbf{ECO (ours)}          & \textbf{93.69} & \textbf{0.981} & 0.009 \\ \hline
\end{tabular}
}

\vspace{-0.5 cm}
\captionof{table}{
Comparison between sparse representation methods on PROXD.
\vspace{-1 cm}
}
\label{tab:sparseCoding}

\end{minipage}

\vspace{-0.3cm}
\end{table*}

\begin{figure*}[h]
   \centering
\resizebox{\linewidth}{!}{
\setlength{\tabcolsep}{2pt}
\begin{tabular}{ccccccccc}
&
\shortstack{\includegraphics[width=0.13\linewidth]{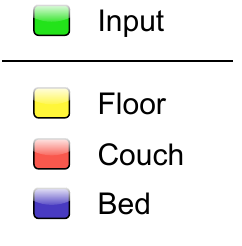}}&
\shortstack{\includegraphics[width=0.13\linewidth]{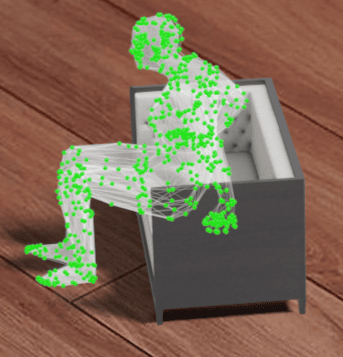}}&
\shortstack{\includegraphics[width=0.13\linewidth]{images/ContactVisCompare/Dense.png}}&
\shortstack{\includegraphics[width=0.13\linewidth]{images/ContactVisCompare/Dense.png}}&
\shortstack{\includegraphics[width=0.13\linewidth]{images/ContactVisCompare/Dense.png}}&
\shortstack{\includegraphics[width=0.13\linewidth]{images/ContactVisCompare/Dense.png}}&
\shortstack{\includegraphics[width=0.13\linewidth]{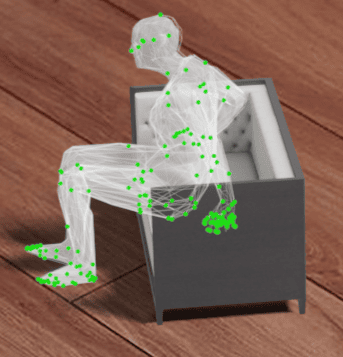}}\\
&
\shortstack{\includegraphics[width=0.13\linewidth]{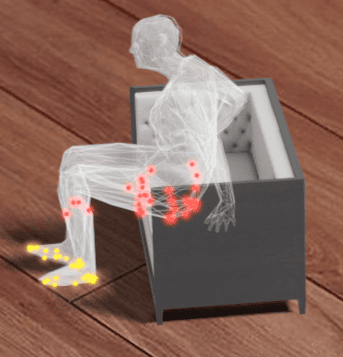}}&
\shortstack{\includegraphics[width=0.13\linewidth]{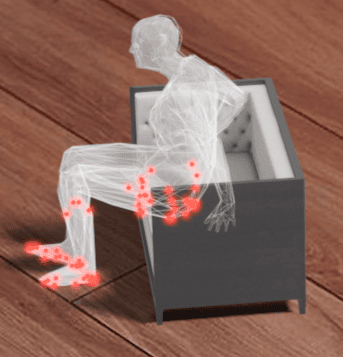}}&
\shortstack{\includegraphics[width=0.13\linewidth]{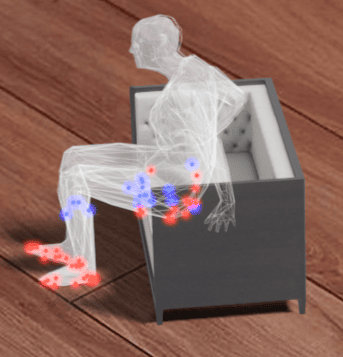}}&
\shortstack{\includegraphics[width=0.13\linewidth]{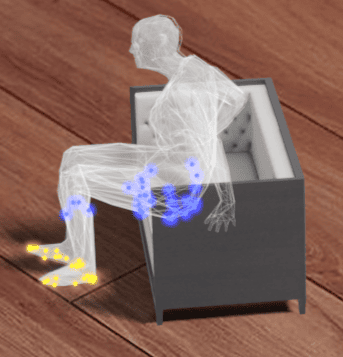}}&
\shortstack{\includegraphics[width=0.13\linewidth]{images/ContactVisCompare/PIAL-Net.png}}&
\shortstack{\includegraphics[width=0.13\linewidth]{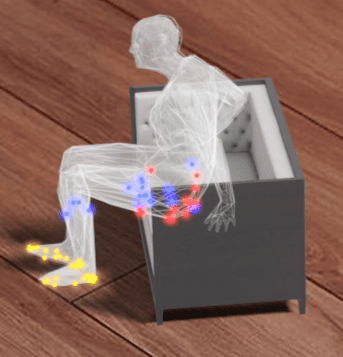}}&
\shortstack{\includegraphics[width=0.13\linewidth]{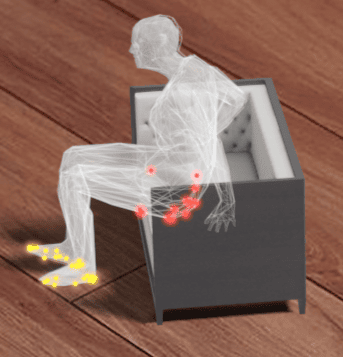}}\\
[-5pt]
&\shortstack{\tiny GT}& 
\shortstack{\tiny LSTM~\cite{greff2016lstm}}&
\shortstack{\tiny POSA~\cite{hassan2021populatingPOSA}}&
\shortstack{\tiny CFormer~\cite{ye2022scene}}&
\shortstack{\tiny PIAL-Net~\cite{luo2023leverage}}&
\shortstack{\tiny HOT~\cite{chen2023detecting}}&
\shortstack{\tiny Ours}
\end{tabular}
}
\vspace{-0.2cm}
    \caption{\textbf{Contact prediction visualization}. The first row includes the contact notation and inputs while the second row shows prediction results. 
    We can see that LSTM~\cite{greff2016lstm} and POSA~\cite{hassan2021populatingPOSA}  show the mismatch between the \texttt{Floor} and the \texttt{Couch}; ContactFormer~\cite{ye2022scene} and HOT~\cite{chen2023detecting} cannot differentiate between \texttt{Couch} and \texttt{Bed}, while our method shows reasonable predictions.
    }
    \label{fig:SOTAVis}
\end{figure*}

\subsection{Contact Prediction Results}
\textbf{Datasets.} We use PROXD~\cite{hassan2019resolvingPROX}, GIMO~\cite{zheng2022gimo}, and BEHAVE~\cite{bhatnagar2022behave} datasets for contact prediction. In all datasets, human bodies are modeled by SMPL-X format~\cite{SMPL-X:2019}. In the PROXD dataset, the contact labels are from PROX-E dataset~\cite{zhang2020generating}. 



\textbf{Evaluation Metrics}. As in~\cite{ye2022scene}, the Reconstruction Accuracy and Consistency Score are used for comparison. We also compare the inference time (second/sample) of all methods on the same NVIDIA Tesla V100 GPU.


\textbf{Baselines}. We compare our method with recent works, including POSA \cite{hassan2021populatingPOSA}, ContactFormer~\cite{ye2022scene}, multi-layer perceptron predictor or bidirectional LSTM \cite{greff2016lstm}, MIME~\cite{yi2023mime}, PIAL-Net~\cite{luo2023leverage}, TRUMANS~\cite{jiang2024scaling}, Ins-HOI\cite{zhang2025ins} and HOT~\cite{chen2023detecting}. We train our ECO using $K=10$ masks and keep only $\kappa=3$ masks with the highest values of mask score $\bm\alpha$ during inference. 

\textbf{Results.} Table~\ref{tab:sota} shows the comparison between our method and other baselines. This table indicates that our model surpasses all other baselines by a large margin with a reconstruction accuracy of $93.69\%$, and a consistency score of $0.981$. Furthermore, our inference speed is $0.009$ second/sample, which is 12 times faster than the runner-up.


\textbf{Visualization.} Fig.~\ref{fig:SOTAVis} shows the qualitative comparison of contact prediction results with different methods. We can see that our method stands out by achieving accurate contact predictions in both the contact labels and contact locations while other methods show a mismatch on contact points.

\subsection{Scene Synthesis Results}
\textbf{Datasets.} 
We use the PROXD \cite{hassan2019resolvingPROX} and GIMO~\cite{zheng2022gimo} datasets for conducting experiments as in recent works. Note that BEHAVE~\cite{bhatnagar2022behave} dataset cannot be used in the scene synthesis task since this dataset only has contacts with independent objects, not ones synchronized in a scene.

\textbf{Baselines.} 
We compare our method with recent baselines on the scene synthesis domain, including ContactICP~\cite{besl1992method}, PosePrior~\cite{moreno2008pose}, SUMMON~\cite{ye2022scene}, MIME~\cite{yi2023mime}, SceneDiffuser~\cite{huang2023diffusion}, HAISOR~\cite{sun2024haisor},  and INFERACT~\cite{li2024physics}. 


\textbf{Evaluation Metric}. We use non-collision score~\cite{zhang2020generating} as a metric for the scene synthesis task. We also perform a user study to compare different methods.

\begin{figure*}[!t] 
  \centering
  \large
  \vspace{-0.2cm}
\resizebox{\linewidth}{!}{
\setlength{\tabcolsep}{2pt}
\begin{tabular}{cccccccc}

\rotatebox[origin=l]{90}{\hspace{-0.1cm} 
{\begin{tabular}[c]{@{}c@{}}ContactICP
\end{tabular}}} &
\shortstack{\includegraphics[width=0.33\linewidth]{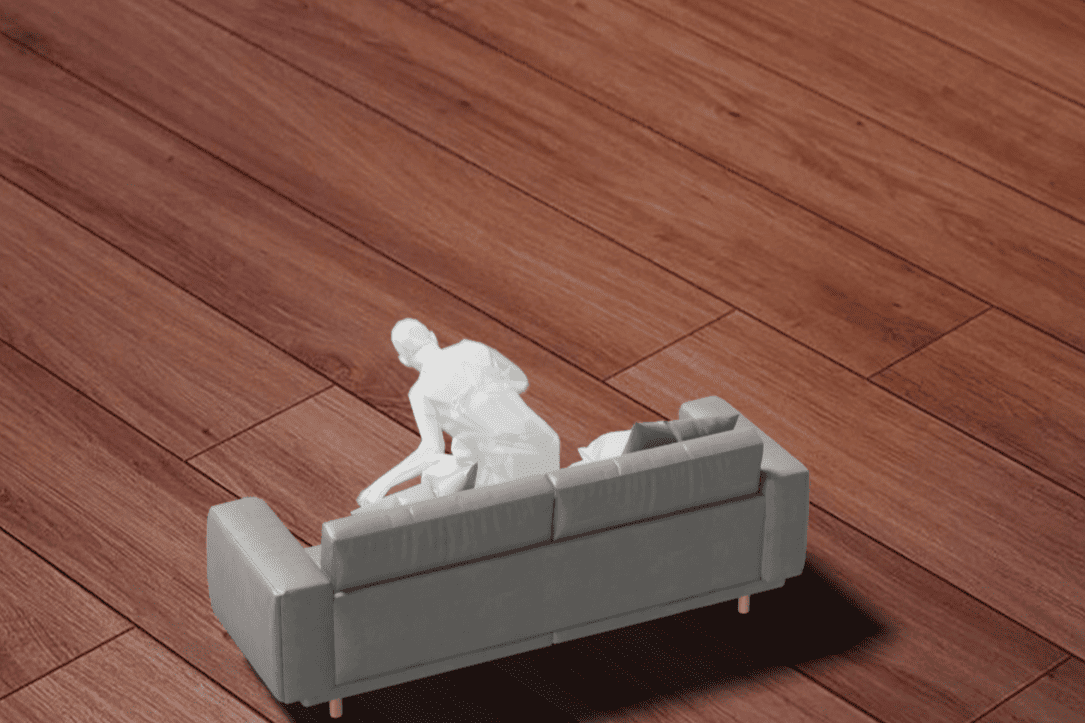}}&
\shortstack{\includegraphics[width=0.33\linewidth]{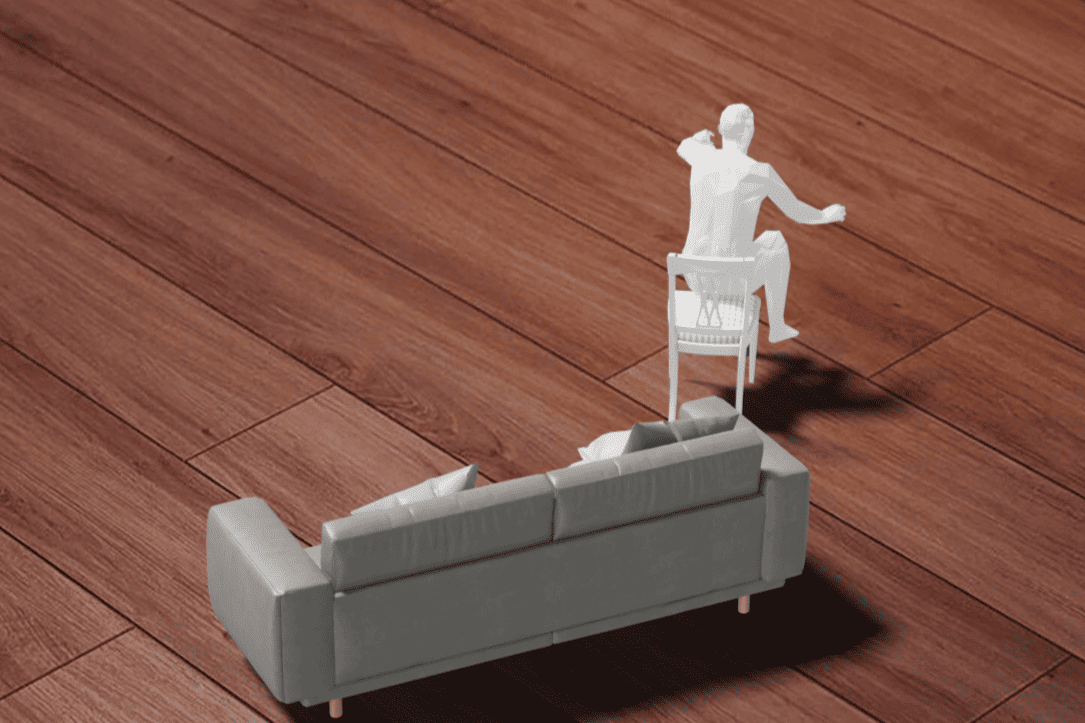}}&
\shortstack{\includegraphics[width=0.33\linewidth]{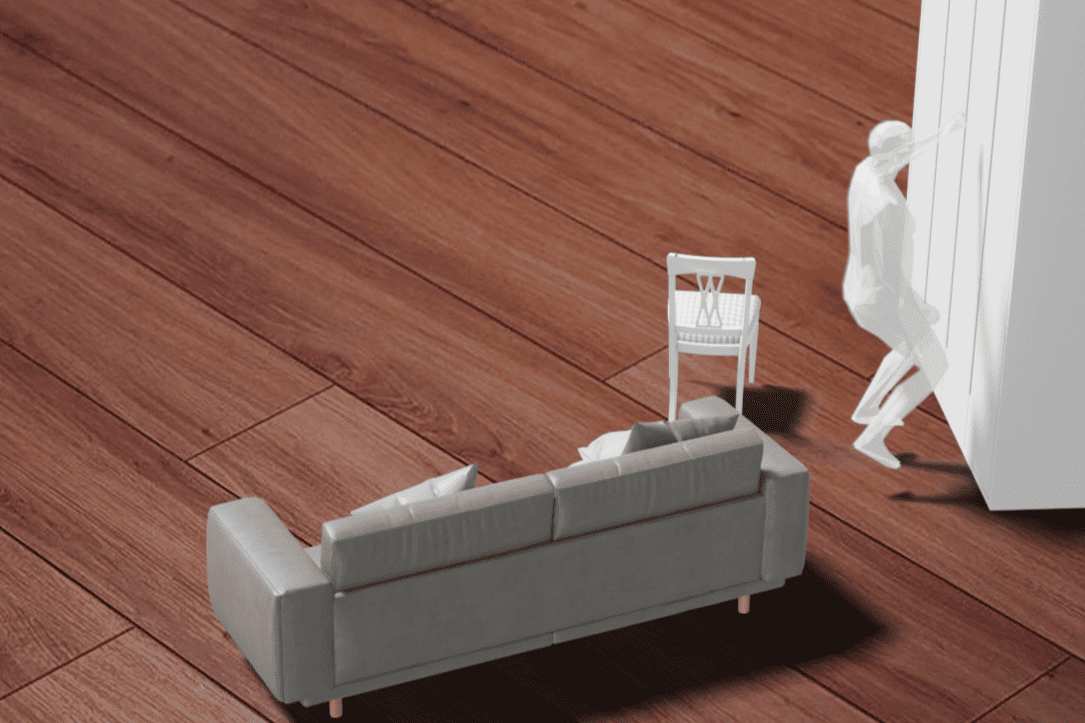}}&
\shortstack{\includegraphics[width=0.33\linewidth]{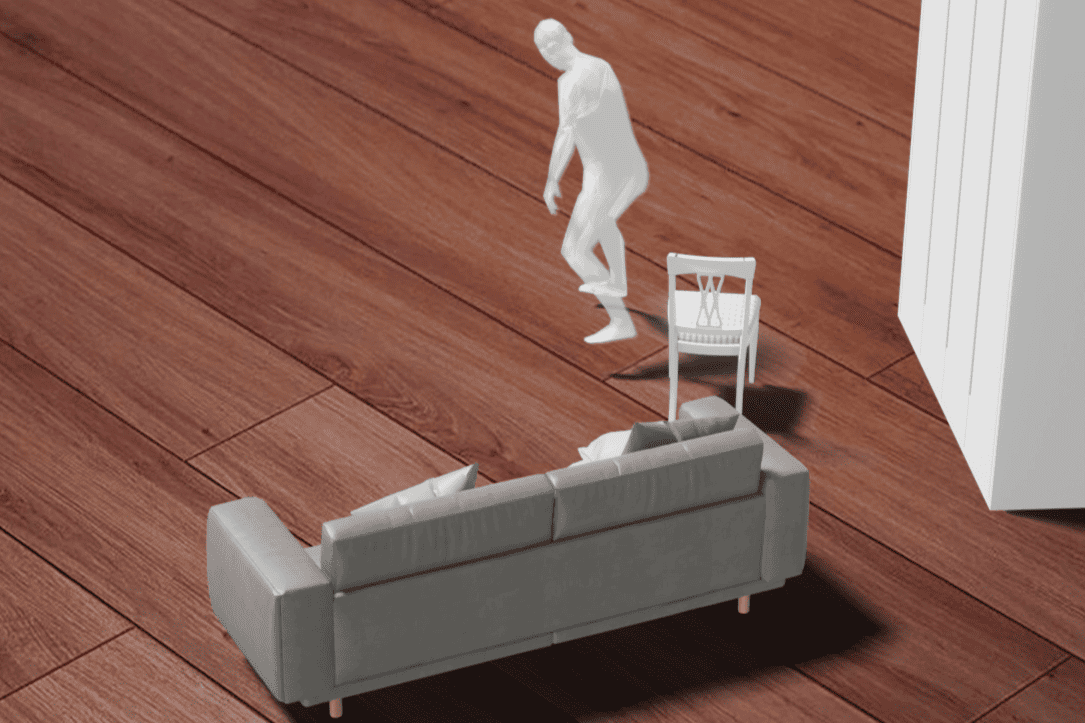}}&
\shortstack{\includegraphics[width=0.33\linewidth]{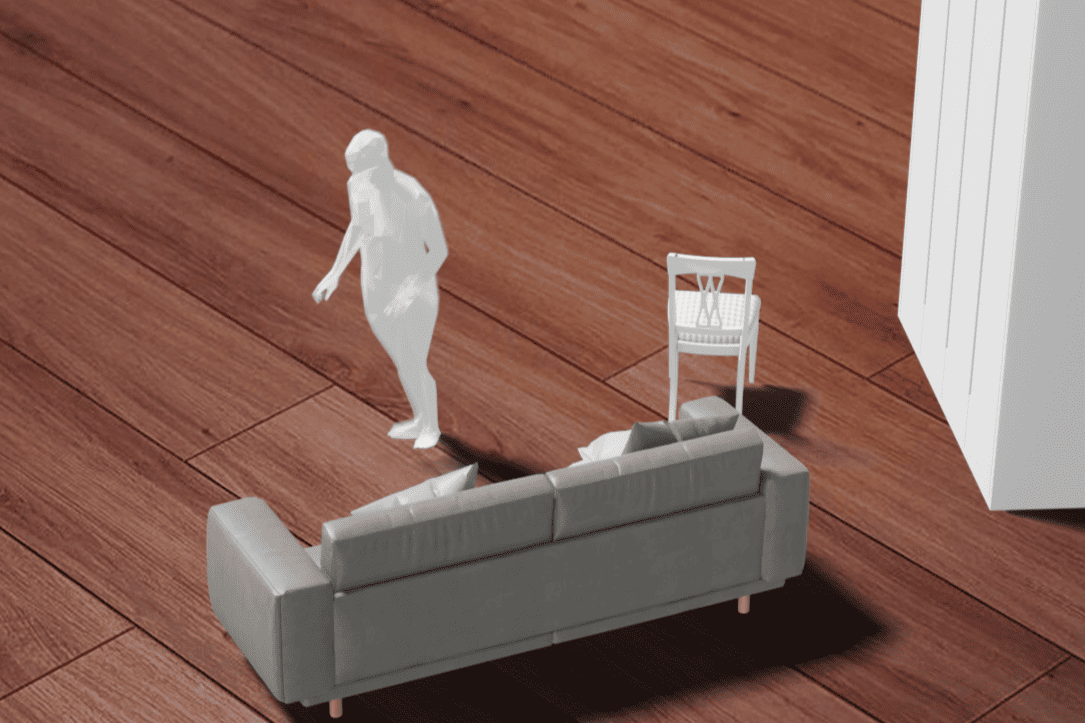}}\\[1pt]


\rotatebox[origin=l]{90}{\hspace{0.2cm} 
{\begin{tabular}[c]{@{}c@{}}SUMMON
\end{tabular}}} &
\shortstack{\includegraphics[width=0.33\linewidth]{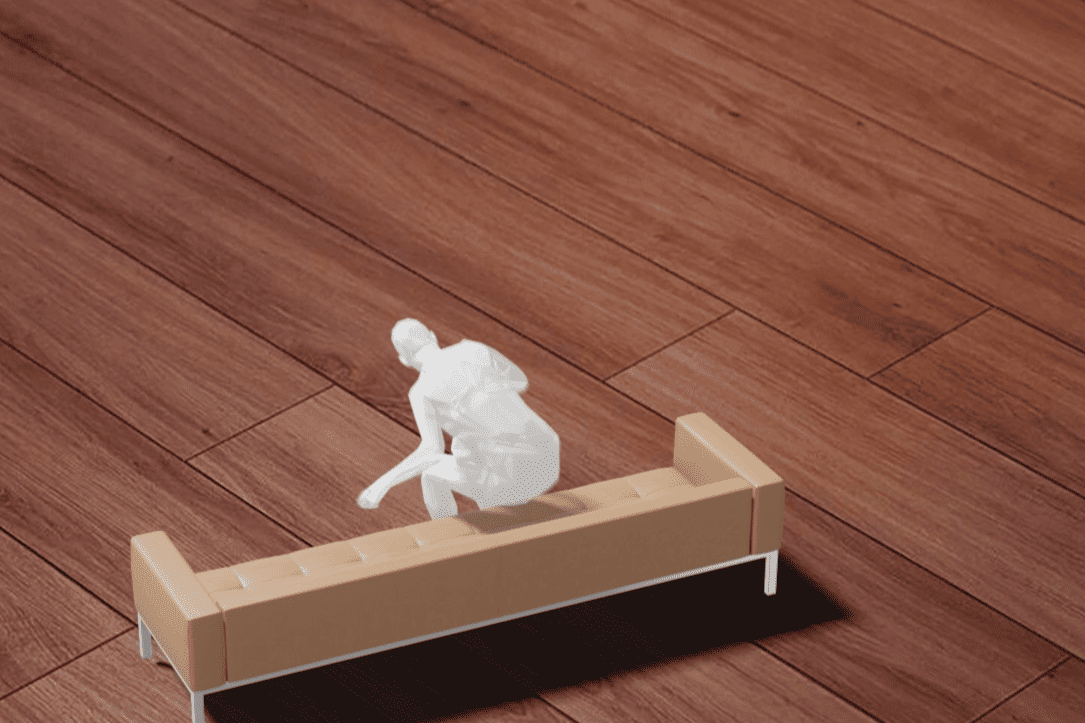}}&
\shortstack{\includegraphics[width=0.33\linewidth]{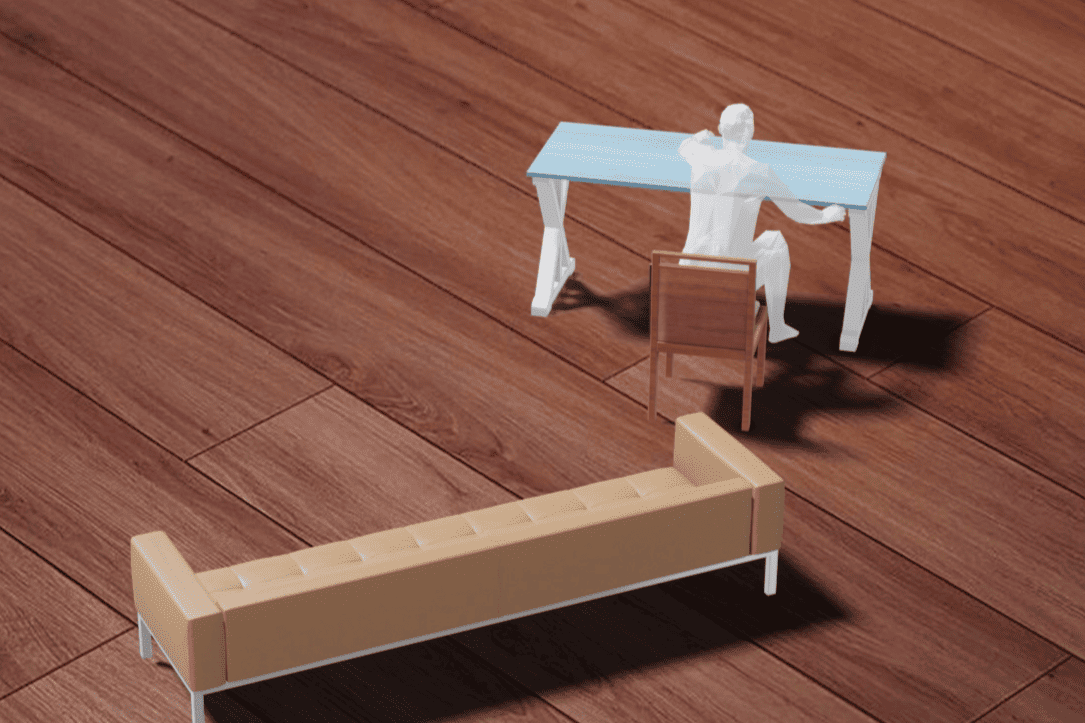}}&
\shortstack{\includegraphics[width=0.33\linewidth]{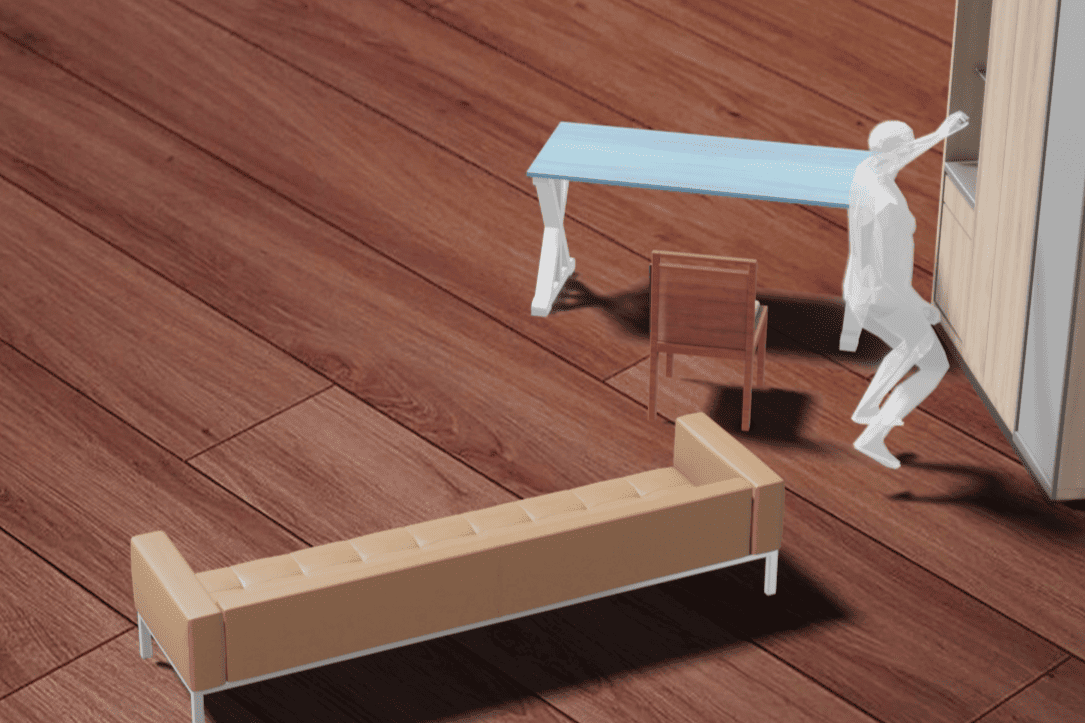}}&
\shortstack{\includegraphics[width=0.33\linewidth]{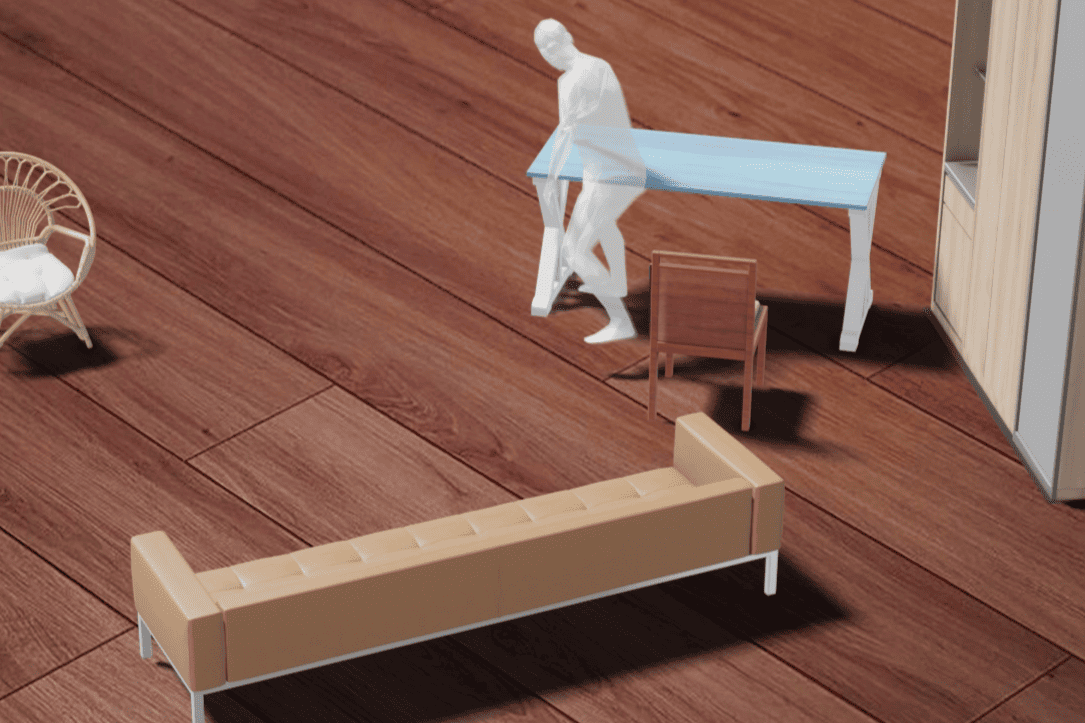}}&
\shortstack{\includegraphics[width=0.33\linewidth]{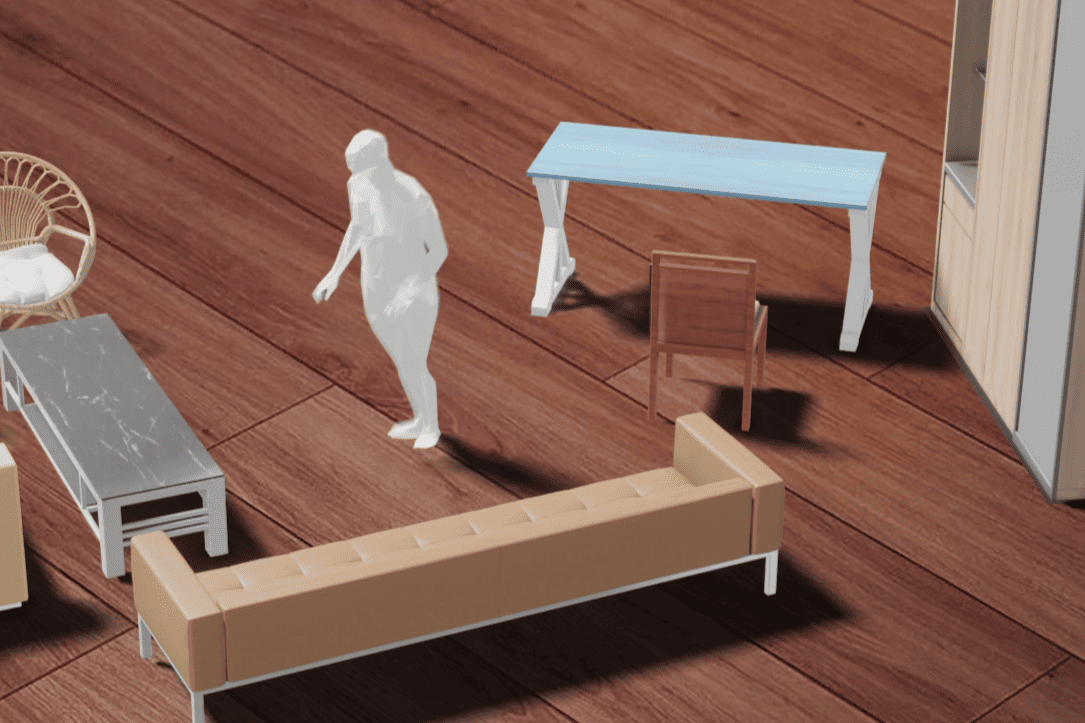}}\\[1pt]

\rotatebox[origin=l]{90}{\hspace{-0cm} 
{\begin{tabular}[c]{@{}c@{}}SceneDiffuser
\end{tabular}}} &
\shortstack{\includegraphics[width=0.33\linewidth]{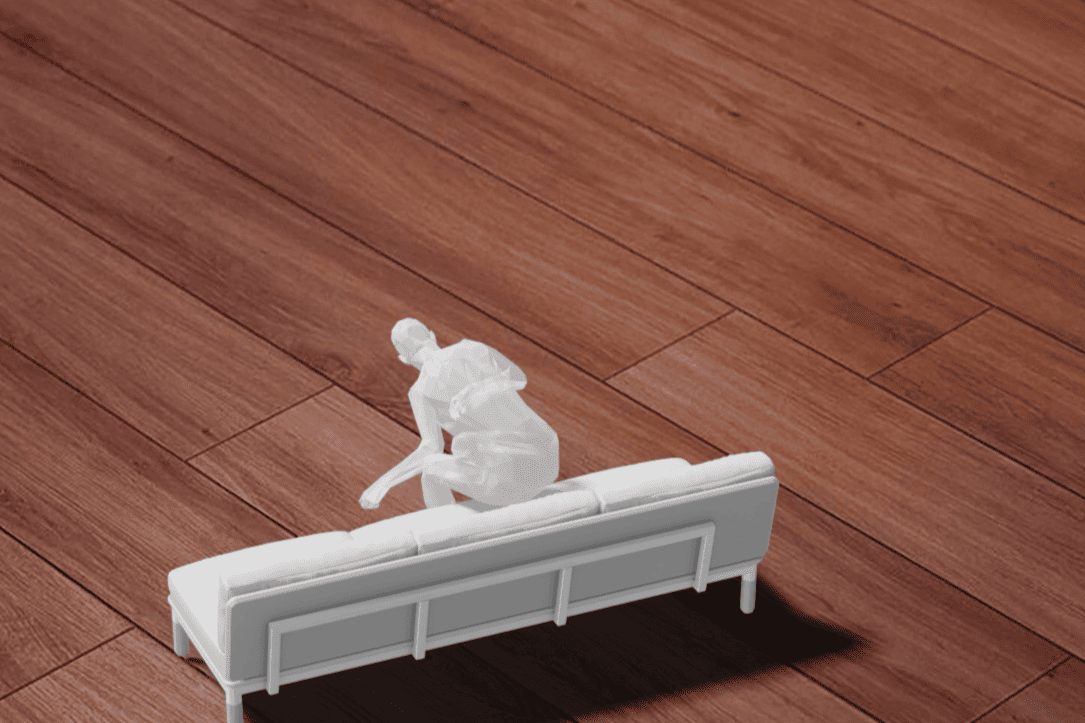}}&
\shortstack{\includegraphics[width=0.33\linewidth]{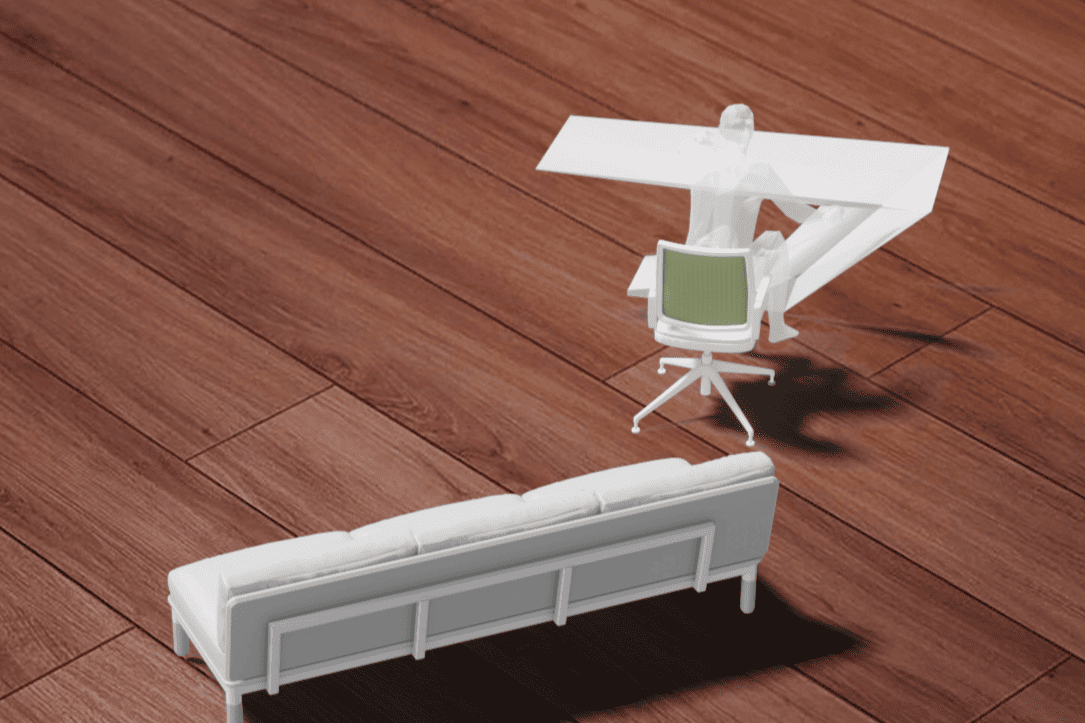}}&
\shortstack{\includegraphics[width=0.33\linewidth]{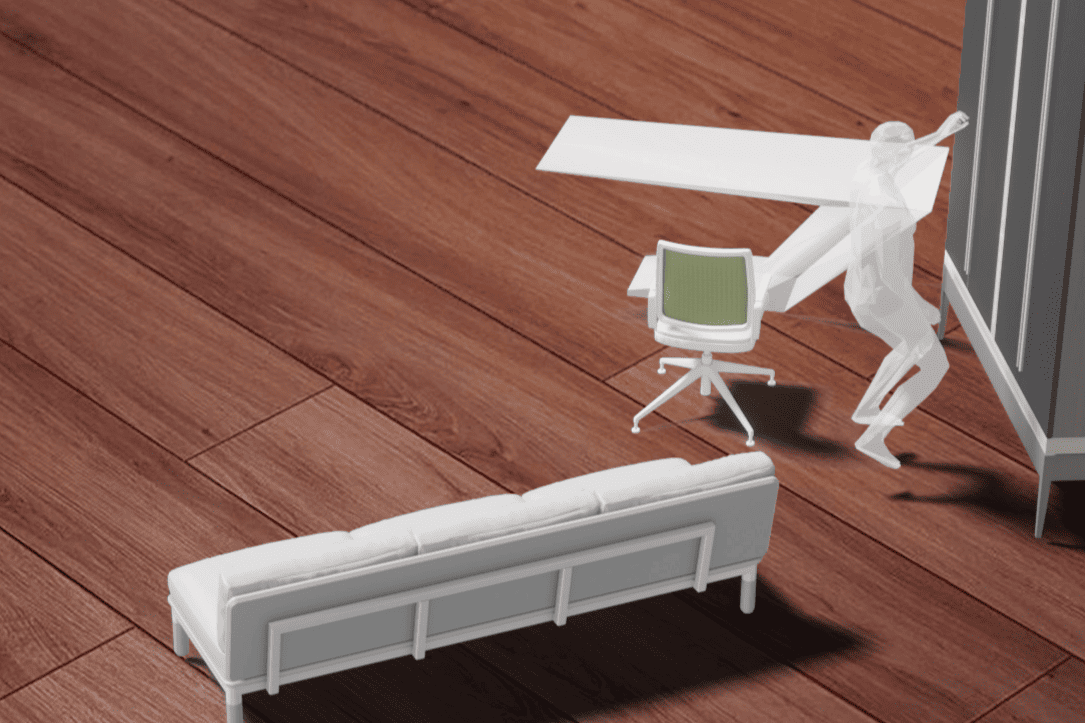}}&
\shortstack{\includegraphics[width=0.33\linewidth]{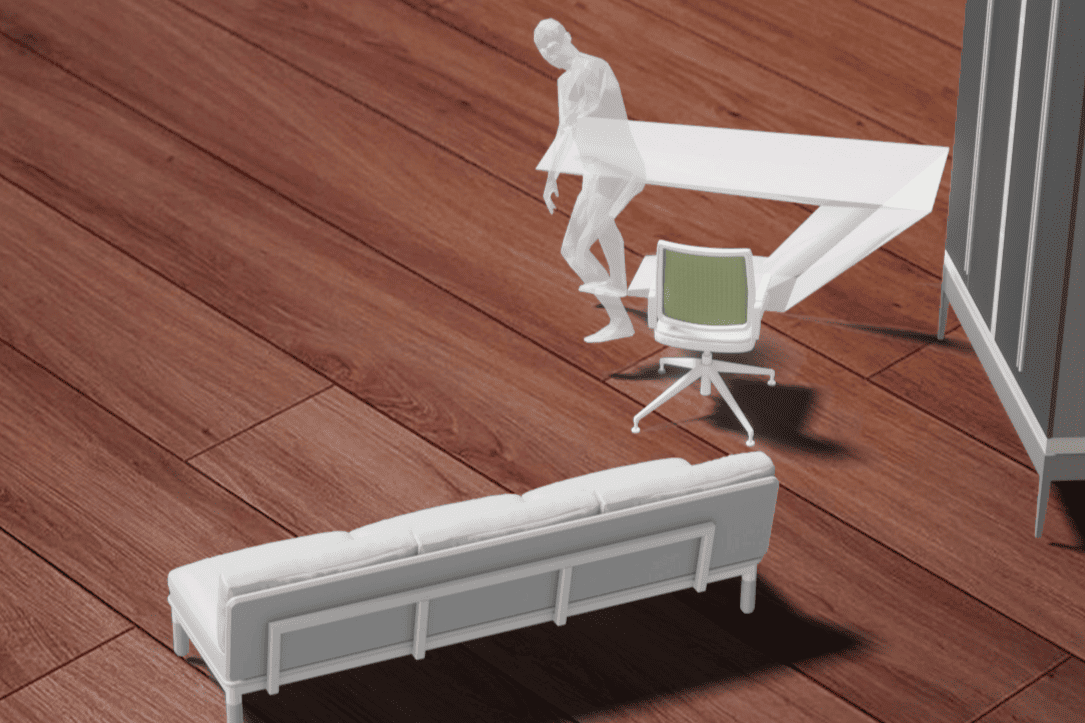}}&
\shortstack{\includegraphics[width=0.33\linewidth]{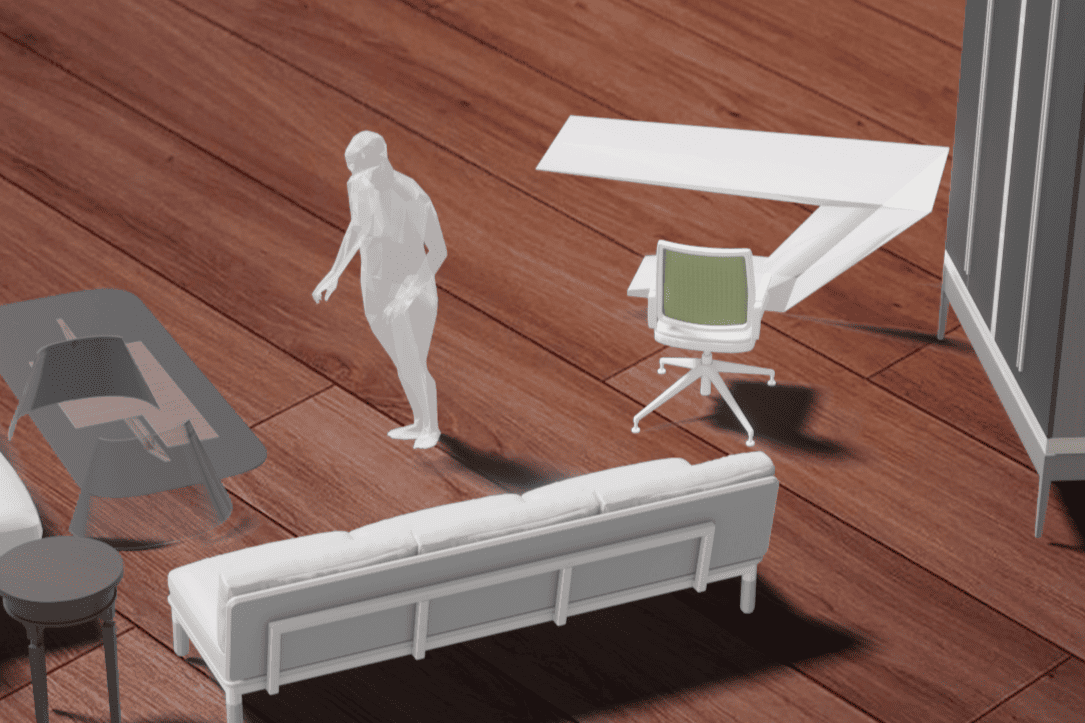}}\\[1pt]

\rotatebox[origin=l]{90}{\hspace{0.2cm} \textbf{\begin{tabular}[c]{@{}c@{}}\hspace{0.3cm} Ours\end{tabular} }} &
\shortstack{\includegraphics[width=0.33\linewidth]{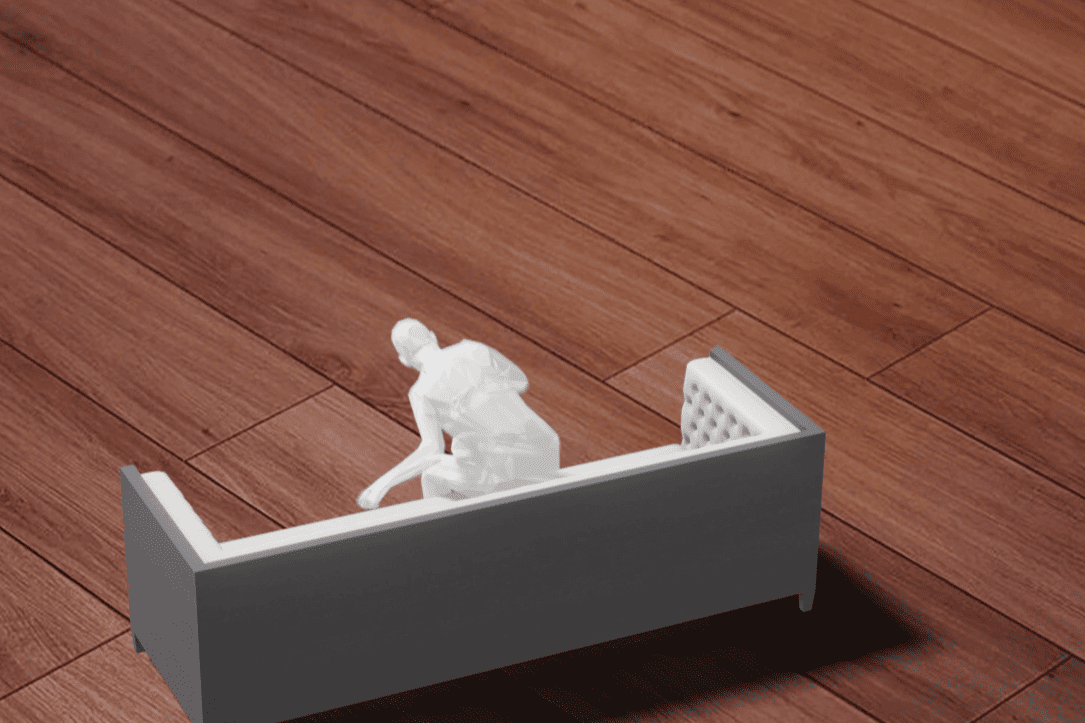}}&
\shortstack{\includegraphics[width=0.33\linewidth]{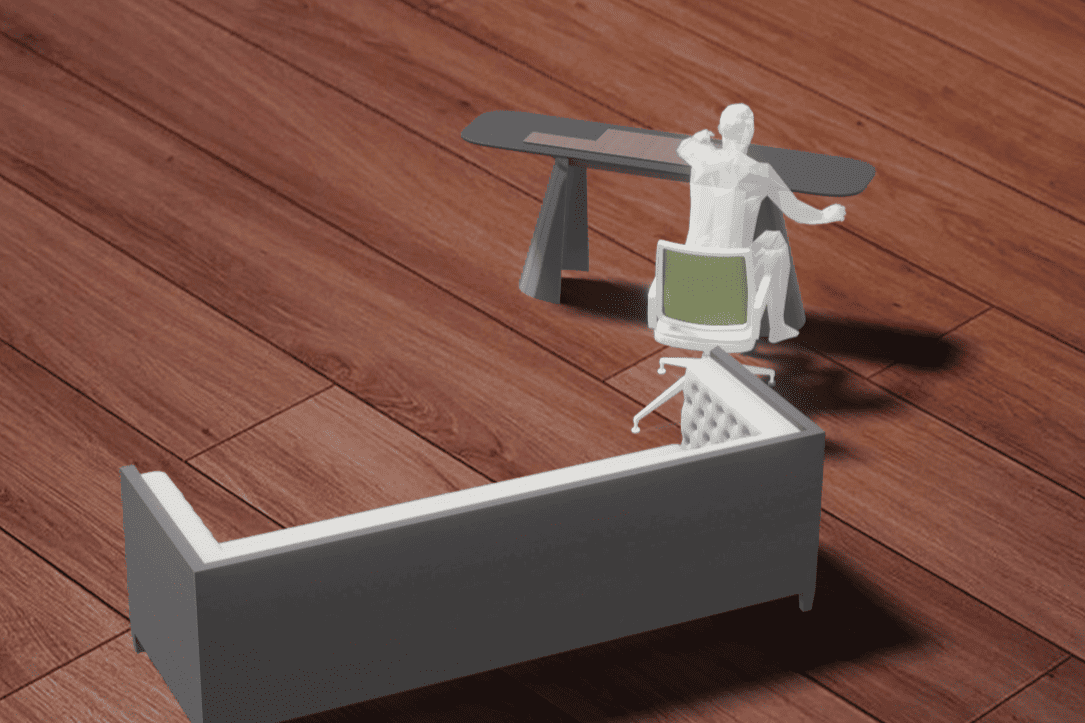}}&
\shortstack{\includegraphics[width=0.33\linewidth]{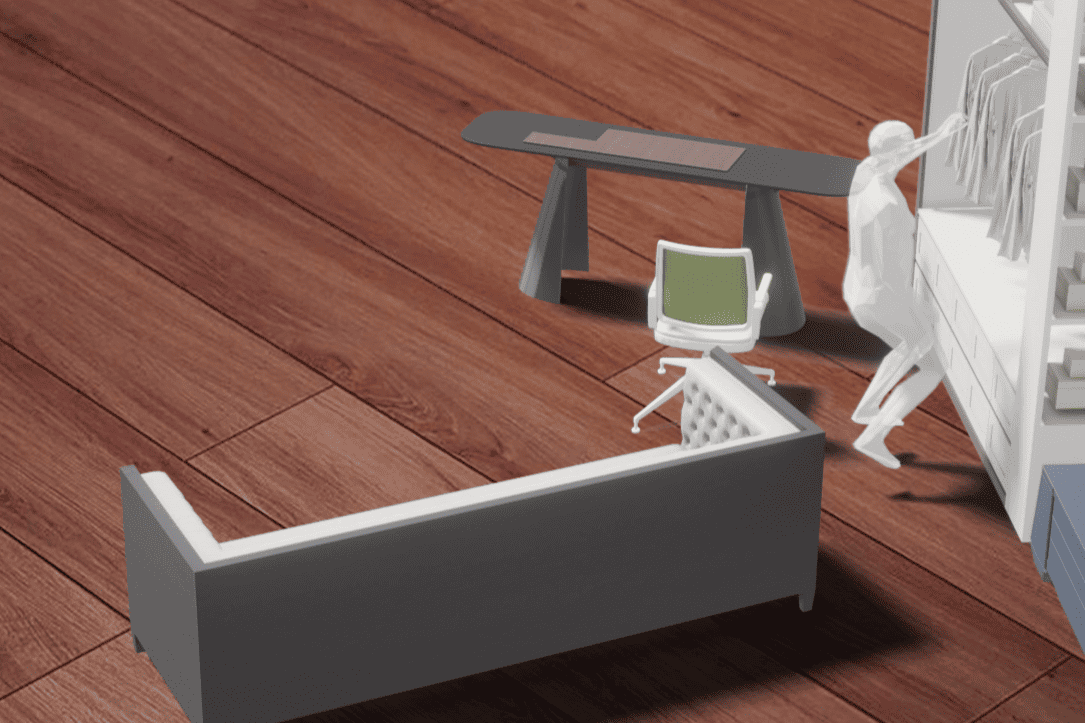}}&
\shortstack{\includegraphics[width=0.33\linewidth]{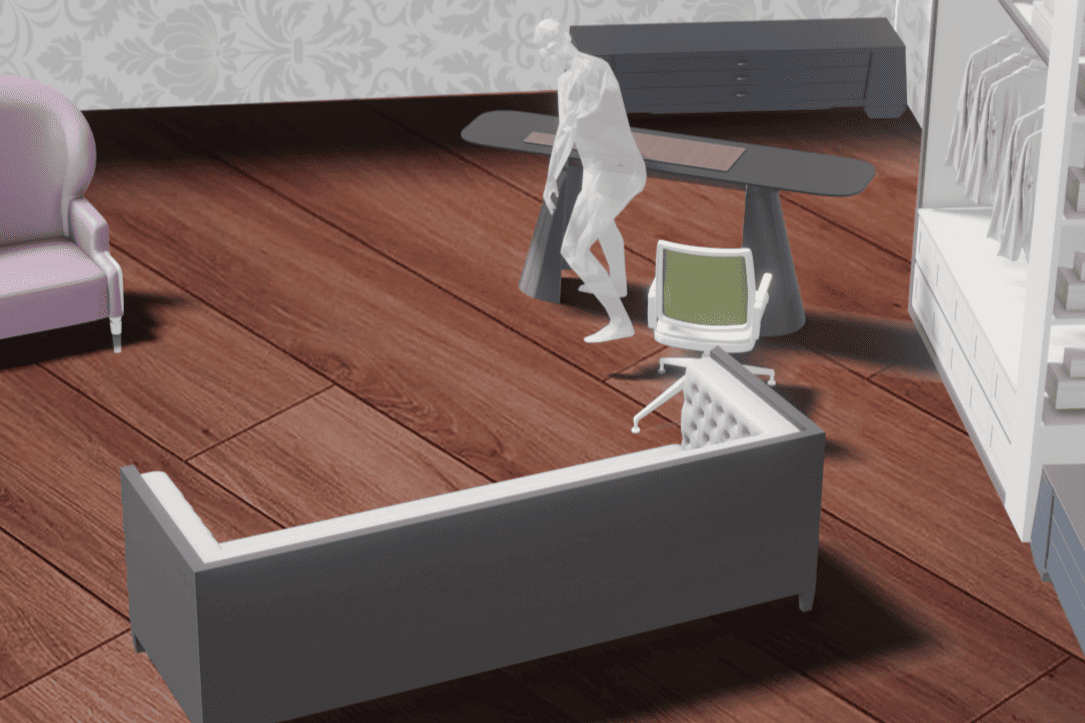}}&
\shortstack{\includegraphics[width=0.33\linewidth]{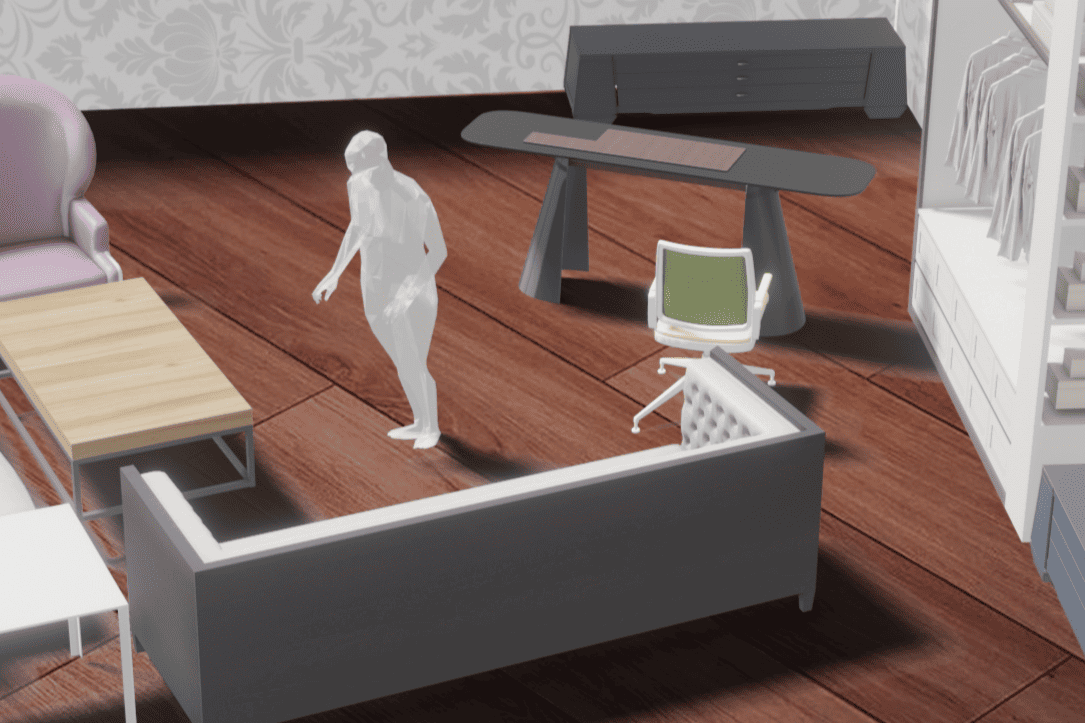}}\\[1pt]

\end{tabular}
}
\vspace{-0.2 cm}
    \caption{\textbf{Scene synthesis visualization between different methods}. Our method efficiently utilizes predicted contacts to produce more reasonable and comprehensive scenes. In contrast, ContactICP struggles to utilize predicted contacts for generating corresponding objects, and SUMMON only supports object positions and orientations without creating a complete scene.
    \vspace{-0.5cm}
    }
    \label{fig:SceneSynthesis}
\end{figure*}

\textbf{Results.} Table~\ref{tab:SceneSynCompare} and Fig.~\ref{fig:SceneSynthesis} show comparisons between scene synthesis results. 
Recent works such as SUMMON~\cite{ye2022scene}, MIME~\cite{yi2023mime}, and SceneDiffuser~\cite{huang2023diffusion} show high reconstruction accuracy, outperforming PosePriors on both datasets. However, our method surpasses all techniques with a clear margin. 

\begin{figure*}[t]
\centering
\RawFloats

\begin{minipage}[t]{0.51\textwidth}
  \centering
  \includegraphics[width= 6.3cm, height  = 6.3 cm]{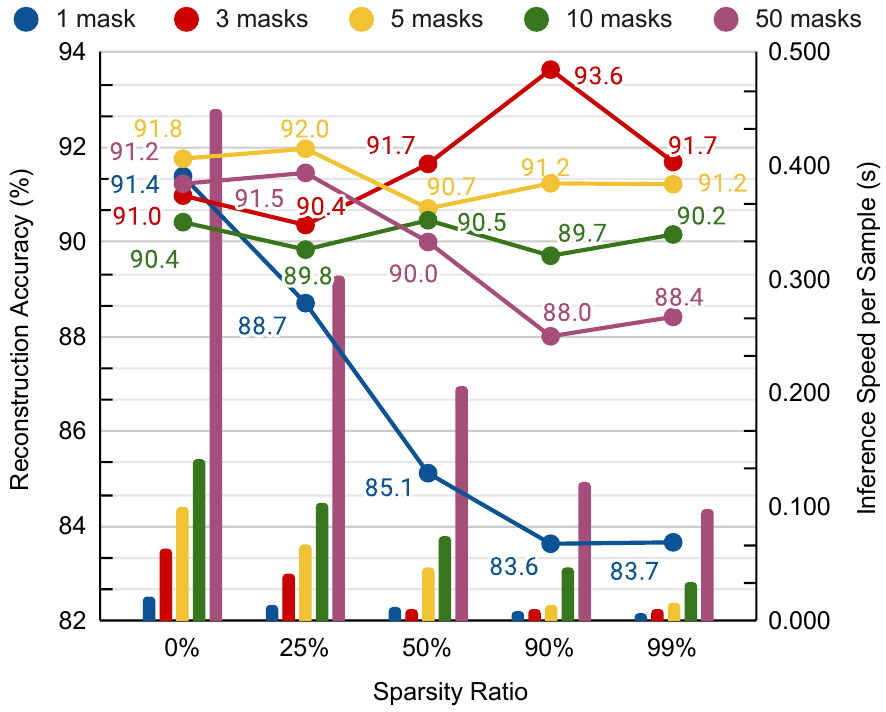}

  \captionof{figure}{
  \textbf{Model Effectiveness with different sparsity ratios and numbers of masks.}
Three sparse masks provide balancing between reconstruction accuracy and inference speed across varying sparsity ratios.
  }
  \vspace{-0.3 cm}
  \label{fig:SparsityVsNumMasksVis}
\end{minipage}
\hfill
\begin{minipage}[t]{0.47\textwidth}
\vspace{-6.3cm}
  \centering

  \resizebox{\linewidth}{!}{
  \setlength{\tabcolsep}{6pt}
  \begin{tabular}{cccc}

  \rotatebox[origin=l]{90}{\tiny Predictions} &
  \includegraphics[width=0.24\linewidth]{images/ContactVisCompare/PIAL-Net.png} &
  \includegraphics[width=0.24\linewidth]{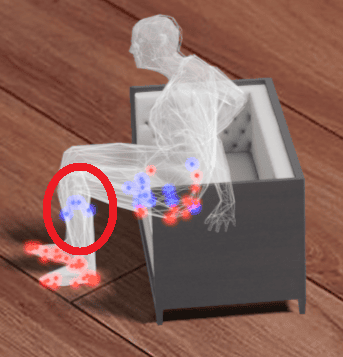} &
  \includegraphics[width=0.24\linewidth]{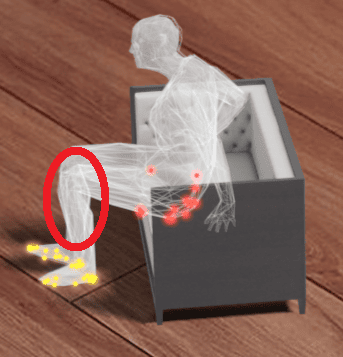}
  \\

  \rotatebox[origin=l]{90}{\tiny Heatmaps} &
  \includegraphics[width=0.24\linewidth]{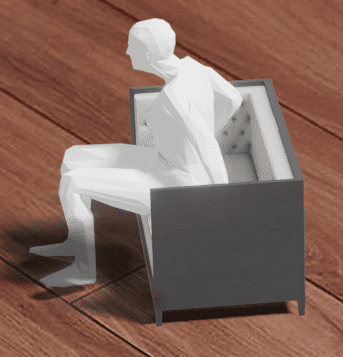} &
  \includegraphics[width=0.24\linewidth]{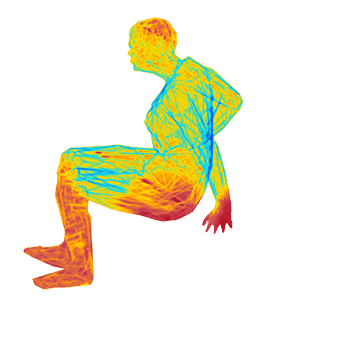} &
  \includegraphics[width=0.24\linewidth]{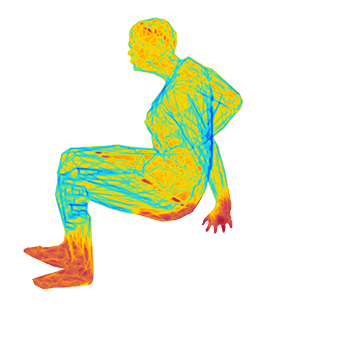}
  \\
\hline \\[-7pt] 
\rotatebox[origin=l]{90}{\tiny Predictions} &
  \includegraphics[width=0.24\linewidth]{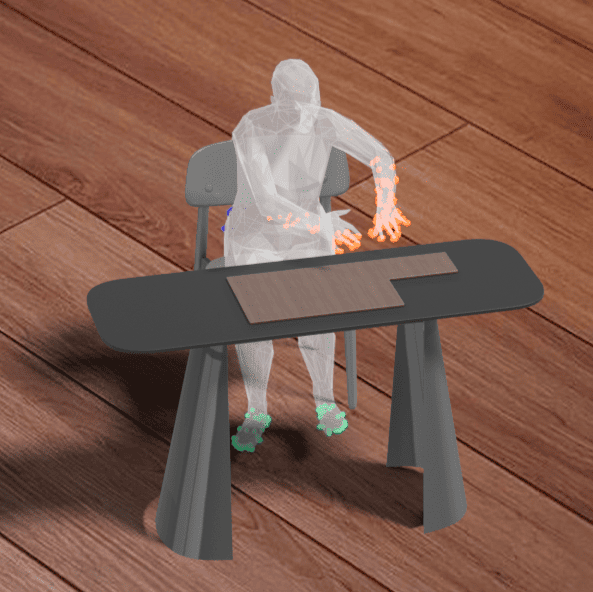} &
  \includegraphics[width=0.24\linewidth]{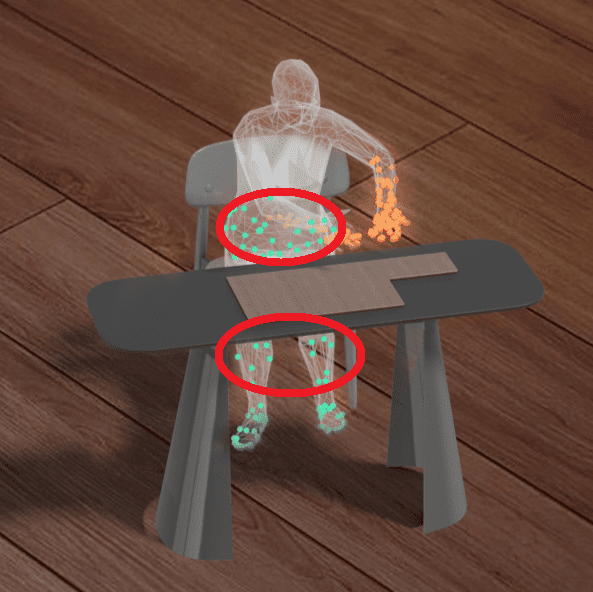} &
  \includegraphics[width=0.24\linewidth]{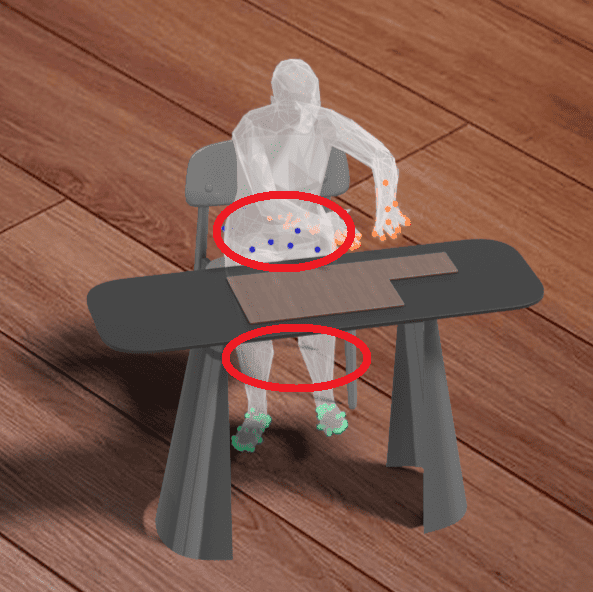}
  \\

  \rotatebox[origin=l]{90}{\tiny Heatmaps} &
  \includegraphics[width=0.24\linewidth]{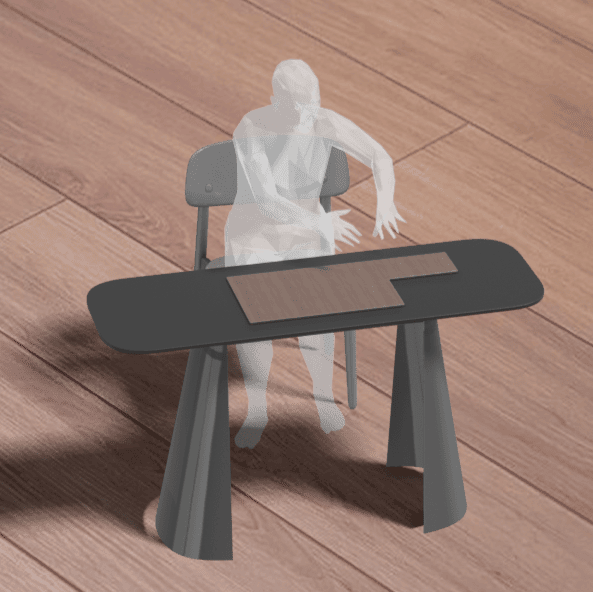} &
  \includegraphics[width=0.24\linewidth]{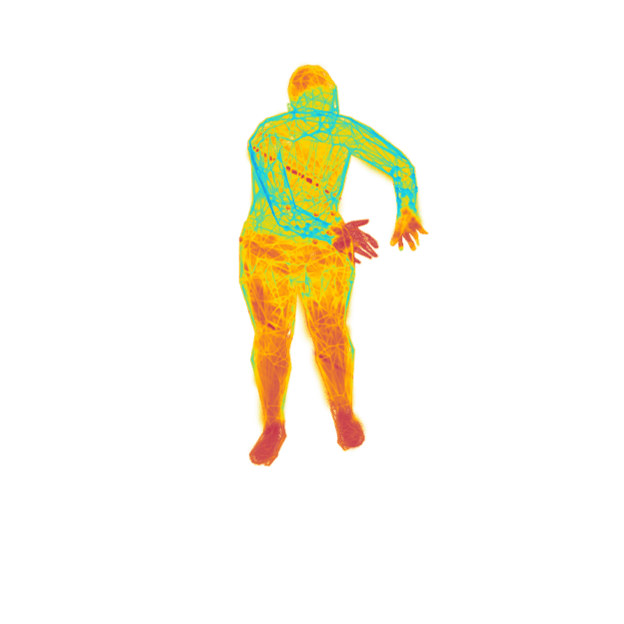} &
  \includegraphics[width=0.24\linewidth]{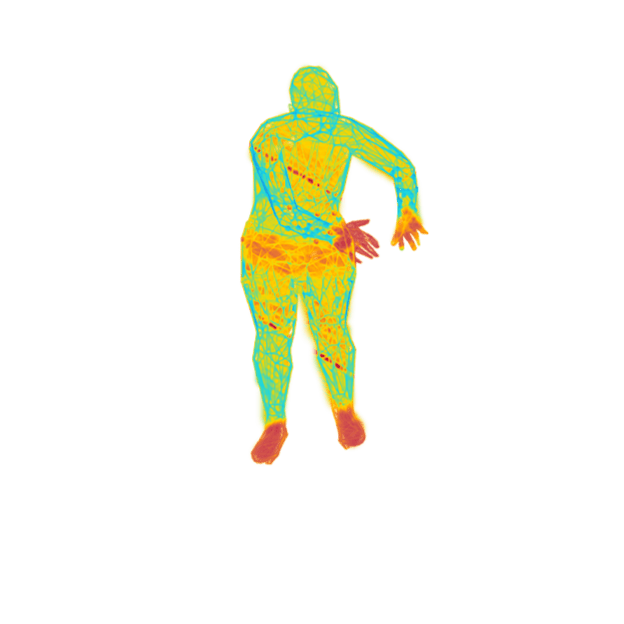}
  \\
  &
  GT &
  POSA &
  Ours
  \\

  \end{tabular}
  }

  \captionof{figure}{
  \textbf{Contact heatmaps visualization}. Red circles indicate that our outputs contain less noisy information and exhibit more concentrated contact regions compared to POSA, resulting in cleaner contact predictions.
  }
  \label{fig:Pattern}
\end{minipage}

\vspace{-0.5cm}
\end{figure*}




\textbf{User Study.}
We conduct a user study with 40 participants from various backgrounds.
Participants are presented with a choice between our method and other models, displayed side by side. Both sets of samples are generated using the PROXD test set. This process is repeated five times for each model and the user scores are from $1$ to $5$. There are two judgment criteria: \textit{(i)} ``Naturalness'' identifies if the position and orientation of facilities are generated properly in the scene and matched with the human poses or not, and \textit{(ii)} ``Non-Collision'' shows if the generated object collides with human motions. The results in Fig.~\ref{fig:UserStudy} show that our method is preferred over other models. 

\subsection{Comparison with Sparse Coding Methods}
\textbf{Baselines.} We compare our method with four representative sparse representation and sparse computation approaches for the contact prediction task, including Minkowski Engine (ME)~\cite{choy20194dMinkowskiME}, EsCoin~\cite{escoin}, pSConv~\cite{spconv}, and 1-D Blocking~\cite{flattened_conv}. For a fair comparison, all methods are integrated into the same contact prediction framework and evaluated under identical experimental settings.

\textbf{Implementation.} We adopt POSA~\cite{hassan2021populatingPOSA} as the backbone architecture for all contact prediction experiments. Following prior work, the input consists of human-scene interaction representations derived from the corresponding datasets. To evaluate both effectiveness and efficiency, we report Reconstruction Accuracy (\%), Consistency Score, and inference speed (seconds/sample). All experiments are conducted using the same hardware configuration and evaluation protocol to ensure fair comparison across methods.

\textbf{Results.} Table \ref{tab:sparseCoding} presents the performance of different sparse representation methods. We can see that our method achieves the highest accuracy compared to all the other sparse coding baselines. For inference speed, our method is only slower than ME~\cite{choy20194dMinkowskiME} ($0.009$ second/sample vs. $0.008$ second/sample) while our accuracy is $10.08\%$ higher.

\begin{table*}[!t]
\centering
\resizebox{1\linewidth}{!}{
\setlength{\tabcolsep}{0.3 em} 
{\renewcommand{\arraystretch}{1.2}
\begin{tabular}{l|c|c|c|c|c|c}
\hline
\multirow{3}{*}{\begin{tabular}[c]{@{}c@{}}\textbf{Test Cases} \end{tabular}} & \multicolumn{6}{c}{\textbf{Criteria}} \\
\cline{2-7}
 & \begin{tabular}[c]{@{}c@{}}\textit{\#Avg. Vertices}\\ \textit{with contacts}$\downarrow$ \end{tabular} & \begin{tabular}[c]{@{}c@{}}\textit{\#Avg. Vertices}\\ \textit{need to predict} $\downarrow$\end{tabular} &  \begin{tabular}[c]{@{}c@{}}\textit{Correct Vertices}\\ \textit{prediction (\%)}$\uparrow$ \end{tabular} & \begin{tabular}[c]{@{}c@{}}\textit{Reconstruction}\\ \textit{Accuracy (\%)}$\uparrow$\end{tabular} & \begin{tabular}[c]{@{}c@{}}\textit{Consistency}\\ \textit{Score}$\uparrow$\end{tabular} &  \begin{tabular}[c]{@{}c@{}}\textit{Inference Speed}\\ \textit{(s/sample)}$\downarrow$ \end{tabular} \\
\hline
\multirow{1}{*}{\begin{tabular}[c]{@{}c@{}}Original Input \end{tabular}} & 121 & 655 & 90.31 & 91.12 &0.882 & 0.28 \\\hline
\multirow{1}{*}{\begin{tabular}[c]{@{}c@{}}Keep all 50 masks \end{tabular}} & 107 \improvecolor{($\downarrow$$\times$ 1.13)}  & 603 \improvecolor{($\downarrow$$\times$ 1.08)} & 88.73 \color[HTML]{d036ff}{(- 1.58)} & 89.46 \color[HTML]{d036ff}{(- 1.66)}   & 0.935 \improvecolor{(+ 0.053)} & 0.451 \color[HTML]{d036ff}{($\uparrow$$\times$ 1.61)} \\\hline
\multirow{1}{*}{\begin{tabular}[c]{@{}c@{}}Keep only 01 mask \end{tabular}} & 12 \improvecolor{($\downarrow$$\times$ 10.08)} & 66 \improvecolor{($\downarrow$$\times$ 9.92)} & 54.67 \color[HTML]{d036ff}{(- 35.64)}  & 83.61 \color[HTML]{d036ff}{(- 7.51)} &0.763 \color[HTML]{d036ff}{(- 0.119)}  &\textbf{ 0.008}  \improvecolor{($\downarrow$$\times$ 35.0)}\\
\rowcolor[HTML]{EFEFEF}\multirow{1}{*}{\begin{tabular}[c]{@{}c@{}}Keep only 03 masks\end{tabular}} & 41 \improvecolor{($\downarrow$$\times$ 2.95)} & 66 \improvecolor{($\downarrow$$\times$ 9.92)}  &  \textbf{95.65} \improvecolor{(+ 5.34)} & \textbf{93.69} \improvecolor{(+ 2.57)} &0.981 \improvecolor{(+ 0.099)}  & 0.009 \improvecolor{($\downarrow$$\times$ 31.1)}\\
\multirow{1}{*}{\begin{tabular}[c]{@{}c@{}}Keep only 10 masks \end{tabular}} & 48 \improvecolor{($\downarrow$$\times$ 2.52)}  & 72 \improvecolor{($\downarrow$$\times$ 9.10)} & 92.07 \improvecolor{(+ 1.76)} & 90.80 \color[HTML]{d036ff}{(- 0.32)}    &\textbf{0.989} \improvecolor{(+ 0.107)} & 0.143 \improvecolor{($\downarrow$$\times$ 1.96)} \\
\hline

\end{tabular}
}
}
\vspace{-0.3 cm}
\caption{Redundant information analysis by selecting masks based on mask score $\bm \alpha$. 
\vspace{-0.3 cm}
\label{tab:RedundantInfoTab}
}
\end{table*}

\subsection{Component Analysis} Table~\ref{tab:ComponentAnalysis} below shows some results regarding the contribution of sparse masks and a sparse network. POSA serves as our baseline, and if setups involve masks, three masks are used. It is evident that when we use sparse masks without the decomposition, the number of data points remains unchanged, leading to no improvement in speed and, in fact, a decrease in accuracy due to missing information. Applying a sparse network to original inputs improves speed, but the trade-off for accuracy is noticeable, as discussed in many previous papers. When we apply the decomposition to the original inputs, the differences in speed and accuracy are not significant compared to the original baseline since the based-decomposition method cannot work properly with full dense tensors. 
With our introduced refinement process that works on integrated decomposition, we can preserve the performance of the model but the speed improvement is not guaranteed. Ultimately, when we use decomposed inputs obtained from sparse masks and a sparse network together, the input shape problem can be addressed, achieving optimization in both speed and accuracy.

\begin{table}[!h]
\vspace{-0.2 cm}
\caption{The effectiveness of each component in our method. Results are benchmarked on the PROXD dataset. 
\vspace{-0.5 cm}
}
\vspace{-0.5 cm}
\centering
\resizebox{1.0\linewidth}{!}{
\setlength{\tabcolsep}{0.25 em} 
{\renewcommand{\arraystretch}{1.0}
\begin{tabular}{c|c|c|c|c|c}
\hline
\textbf{\begin{tabular}[c]{@{}c@{}}Network  Type\end{tabular}} & \textbf{Decompose} & \textbf{\begin{tabular}[c]{@{}c@{}} POSA Sparse\\ Masks\end{tabular}} & \textbf{\begin{tabular}[c]{@{}c@{}}Sparse Mask\\ Refinement\end{tabular}} & \textbf{\begin{tabular}[c]{@{}c@{}}Reconstruction \\ Accuracy (\%)\end{tabular}} & \textbf{\begin{tabular}[c]{@{}c@{}}Inference Speed\\ (s//sample)\end{tabular}} \\ \hline
\rowcolor[HTML]{EFEFEF}Dense  Net. &  &  &  & 91.12 & 0.28 \\ \hline
Dense Net. &  & \checkmark &  & 85.72 \color[HTML]{d036ff}{(- 5.4)} & 0.28 \improvecolor{($\downarrow$$\times$ 1.0)}\\ \hline
\rowcolor[HTML]{EFEFEF}Dense Net.& \checkmark &  &  & 91.02 \color[HTML]{d036ff}{(- 0.1)} & 0.27 \improvecolor{($\downarrow$$\times$ 1.04)} \\ \hline
Dense  Net. & \checkmark & \checkmark &  &85.46 \color[HTML]{d036ff}{(- 5.66)} & 0.28 \improvecolor{($\downarrow$$\times$ 1.0)} \\ \hline
\rowcolor[HTML]{EFEFEF}Dense  Net. & \checkmark & \checkmark & \checkmark & 93.27 \improvecolor{(+ 2.15)} & 0.28 \improvecolor{($\downarrow$$\times$ 1.0)} \\ \hline
Sparse  Net. & \checkmark & \checkmark & \checkmark & \textbf{93.69}  \improvecolor{(+ 2.57)} & \textbf{0.01} \improvecolor{($\downarrow$$\times$ 28)} \\ \hline
\end{tabular}
}}

\label{tab:ComponentAnalysis}
\end{table}
\vspace{-0.3 cm}
\subsection{Sparse Mask Analysis}
\textbf{Sparsity ratio and the number of sparse masks.}
Fig.~\ref{fig:SparsityVsNumMasksVis} illustrates the correlation between reconstruction accuracy and inference speed of our method under different values of sparsity ratio and the number of sparse masks $K$. We note that $K=50$ masks are used during training. During inference, we consequently only select $\kappa$ masks based on the value of mask score $\bm \alpha$. We can see that using $\kappa=1$ mask leads to faster model performance, however, this also significantly reduces accuracy due to the loss of input information. In contrast, employing multiple sparse masks helps retain essential information and improves the overall model performance. Overall, Fig.~\ref{fig:SparsityVsNumMasksVis} shows that using $\kappa=3$ masks with $90\%$ sparsity ratio during the inference brings the balance of the accuracy and inference speed.

\textbf{How sparse masks help reduce redundancy and improve the results?} Our sparse masks act as filters to reduce non-useful information in human-scene input. Specifically, they reduce vertices in representations, influencing both inference speed and accuracy. Table~\ref{tab:RedundantInfoTab} shows how sparse masks help eliminate redundant input. The ``Original Input" uses all vertices, while ``Keep only 01 mask" uses only $\kappa=1$ mask at inference. We also evaluate setups with $3$, $10$, and all $50$ masks. Our method is trained with $K=50$ sparse masks, each with $90\%$ sparsity. Masks are kept based on mask score $\bm{\alpha}$ in Section~\ref{subsec:ECO}. As shown in Table~\ref{tab:RedundantInfoTab}, the model using just 1 sparse mask reduces vertex processing requirements by $90\%$, significantly enhancing inference speed but causing a $7.51\%$ accuracy drop compared to the ``Original Input" setup. With $50$ masks, our ECO maintains accuracy but increases inference time since too many masks are used. Using mask score $\bm{\alpha}$, we can remove non-useful masks and retain only $10$ or even $3$ informative masks during inference. 

Fig.~\ref{fig:Pattern} illustrates the ground truth, the contact heatmaps from the baseline POSA using dense tensors, and our ECO approach with sparse masks. Vertices within the highlighted red ellipse are redundant and ignored by ECO, while other methods still consider them, leading to incorrect contact predictions. These results imply that ECO effectively removes redundant input information, improving model performance.

\begin{figure}[ht]
\centering

\begin{tabular}{cc}
\begin{tabular}{@{}c@{}}
\includegraphics[width=0.45\linewidth]{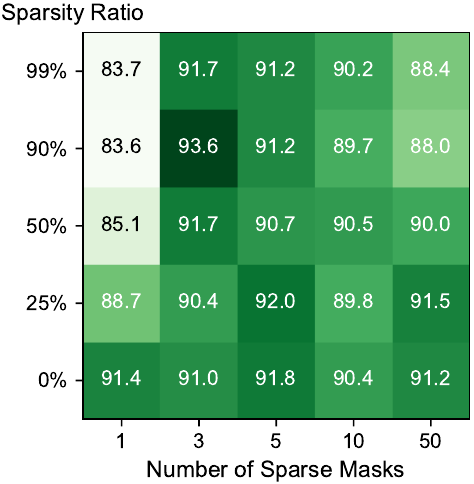} \\
(a) Reconstruction Accuracy (\%)
\end{tabular}
&
\hspace{0.03\linewidth}
\begin{tabular}{@{}c@{}}
\includegraphics[width=0.45\linewidth]{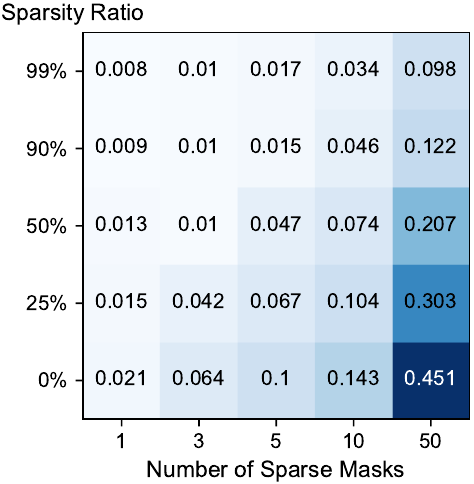} \\
(b) Inference Speed (s/sample)
\end{tabular}
\end{tabular}

\caption{\textbf{Reconstruction Accuracy (\%) and Inference Speed (s/sample) between setups.} The visualization highlights the trade-off between achieving high sparsity in the model, maintaining reconstruction quality, and preserving computational efficiency.
\vspace{-0.5 cm}
}
\label{fig:SparsityVsNumMasks}
\end{figure}

For further clarification, we conduct an extended analysis to examine our proposal.
Figure~\ref{fig:SparsityVsNumMasks} illustrates the results of ECO when we change the number of sparse masks and the sparsity ratio. In particular, Figure~\ref{fig:SparsityVsNumMasks}.a shows the Reconstruction Accuracy, and Figure~\ref{fig:SparsityVsNumMasks}.b demonstrates the corresponding GPU inference time. The results indicate that as the number of masks and the sparsity ratio increase, the inference speed decreases. Additionally, more redundant masks can be established. However, with an appropriate trade-off, state-of-the-art results with efficient inference time can be achieved. We can observe that, with a $90\%$ sparsity ratio and 3 masks, we achieve a state-of-the-art $93.6\%$ accuracy while still maintaining efficient processing during inference ($0.01$ second/sample).

\subsection{Feature Similarity Analysis } 
\begin{figure*}[t]
\centering
\RawFloats

\begin{minipage}[t]{0.42\textwidth}
\centering

\setlength\tabcolsep{5pt}
\renewcommand\arraystretch{0.6}

\begin{tabular}{cc}
\begin{tabular}{@{}c@{}}
\includegraphics[width=0.45\linewidth]{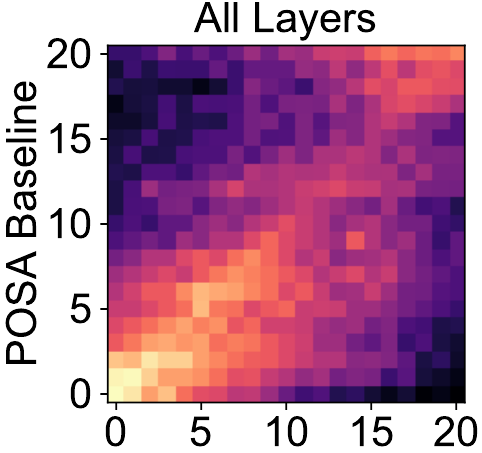} \\
\tiny (a) Keep 1 mask
\end{tabular}
&
\begin{tabular}{@{}c@{}}
\includegraphics[width=0.45\linewidth]{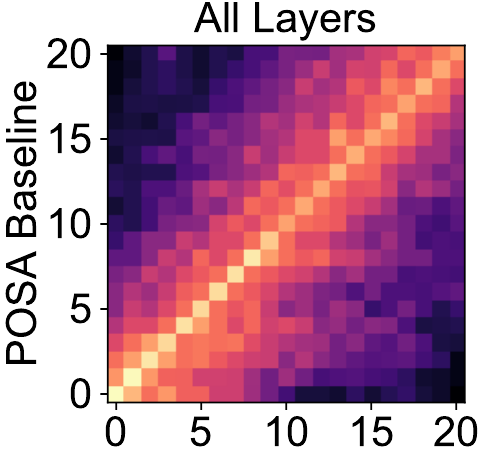} \\
\tiny (b) Keep 3 masks
\end{tabular}
\\\\
\begin{tabular}{@{}c@{}}
\includegraphics[width=0.45\linewidth]{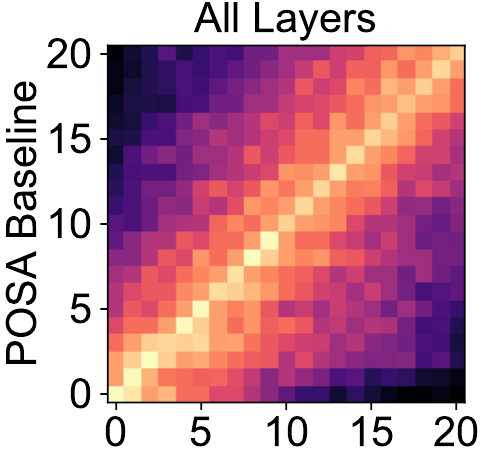} \\
\tiny (c) Keep 5 masks
\end{tabular}
&
\begin{tabular}{@{}c@{}}
\includegraphics[width=0.45\linewidth]{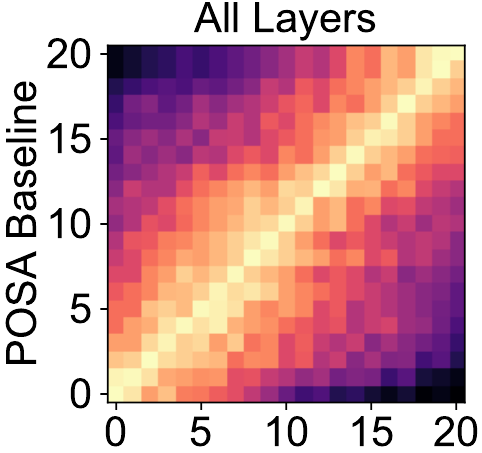} \\
\tiny (d) Keep 10 masks
\end{tabular}
\end{tabular}
\vspace{-0.25cm}
\captionof{figure}{
Similarity between outputs of intermediate layers.
}
\label{fig:NetworkSimilarity}

\end{minipage}
\hfill
\begin{minipage}[t]{0.56\textwidth}
\centering

\setlength\tabcolsep{5pt}
\renewcommand\arraystretch{0.6}

\begin{tabular}{cc}
\begin{tabular}{@{}c@{}}
\includegraphics[width=0.47\linewidth]{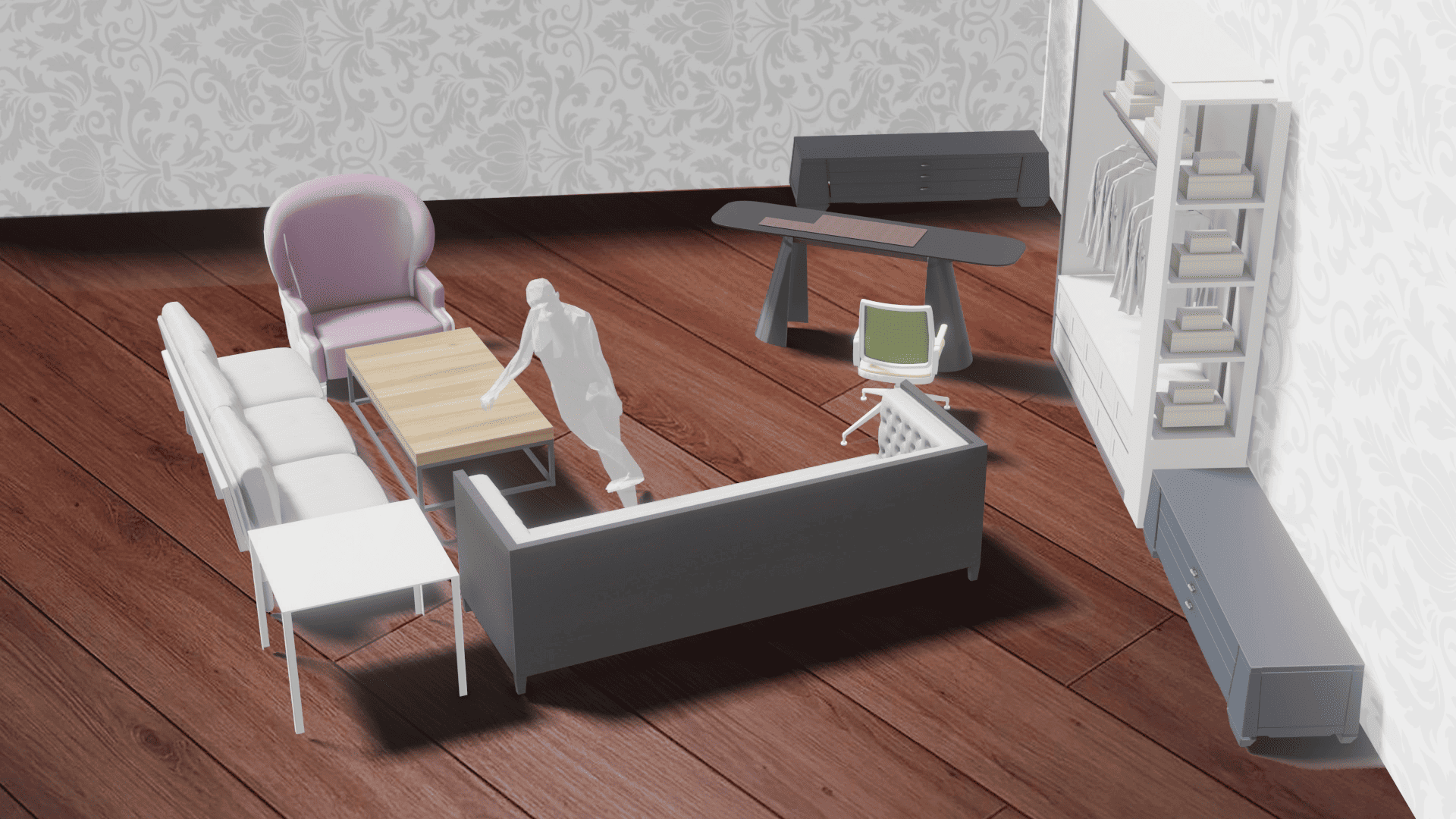} \\
\tiny Good Scene
\end{tabular}
&
\begin{tabular}{@{}c@{}}
\includegraphics[width=0.47\linewidth]{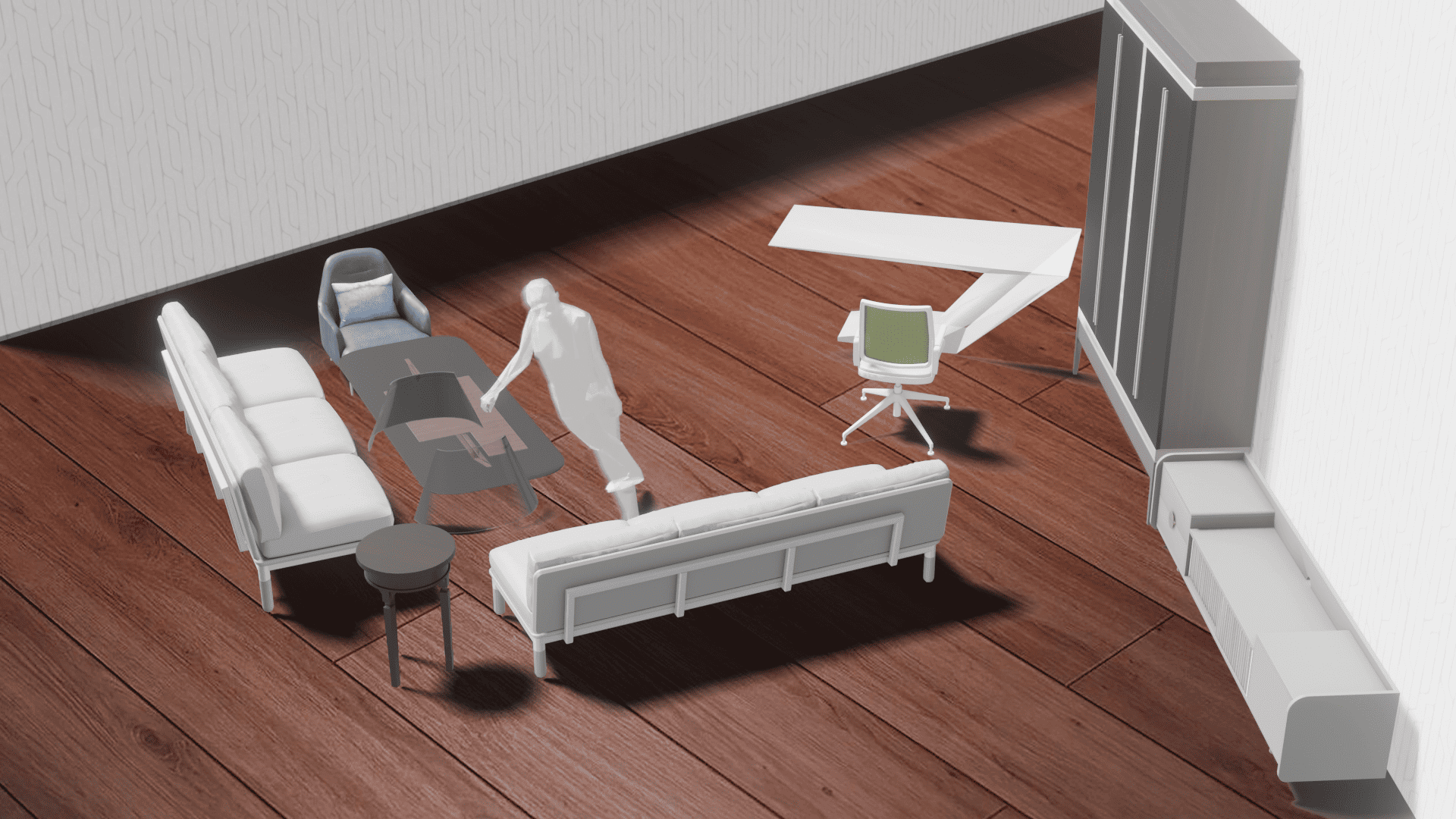} \\
\tiny Wrong Positions (Wall)
\end{tabular}
\\\\[8pt]
\begin{tabular}{@{}c@{}}
\includegraphics[width=0.47\linewidth]{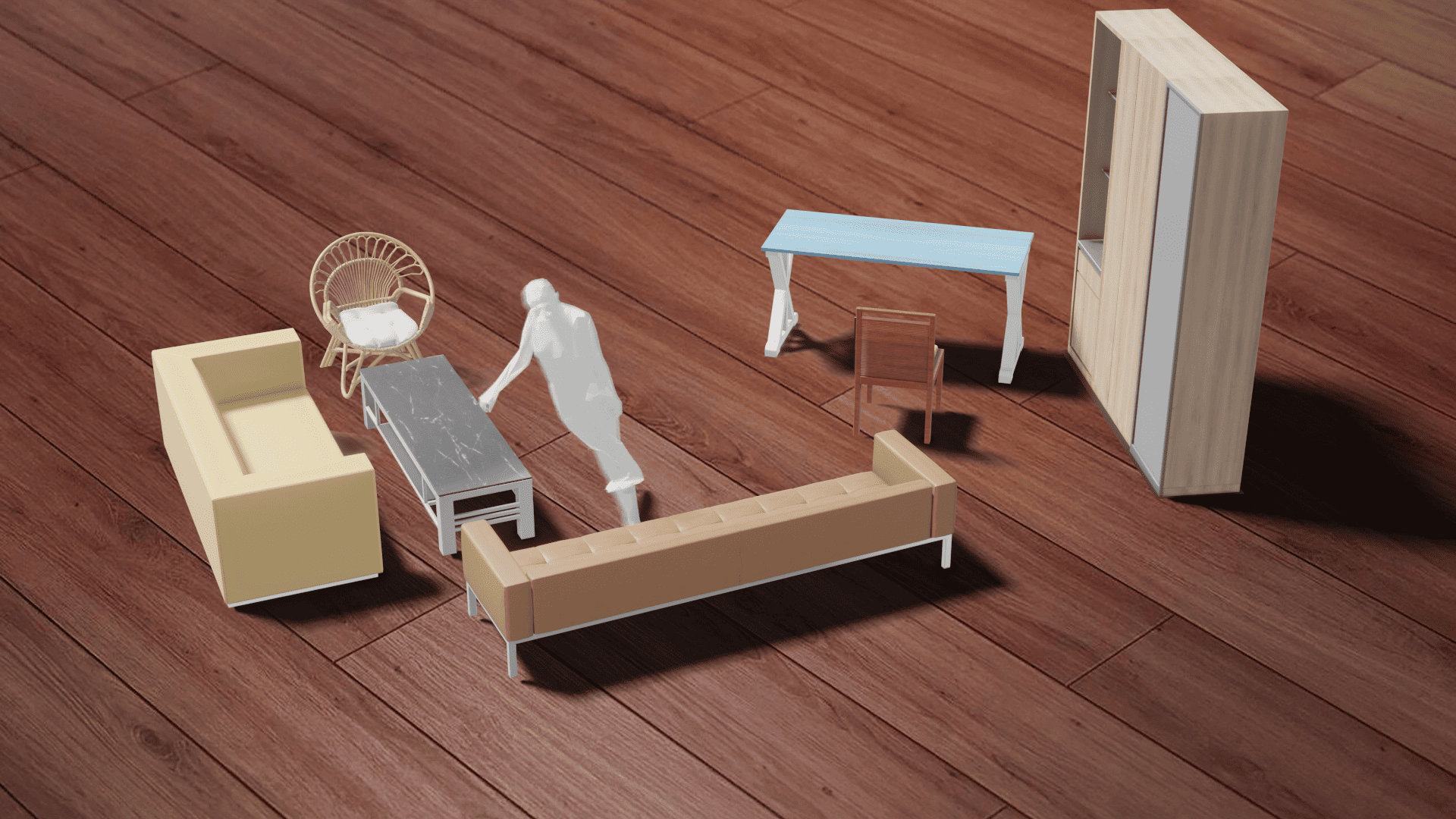} \\
\tiny Miss Objects (Wall)
\end{tabular}
&
\begin{tabular}{@{}c@{}}
\includegraphics[width=0.47\linewidth]{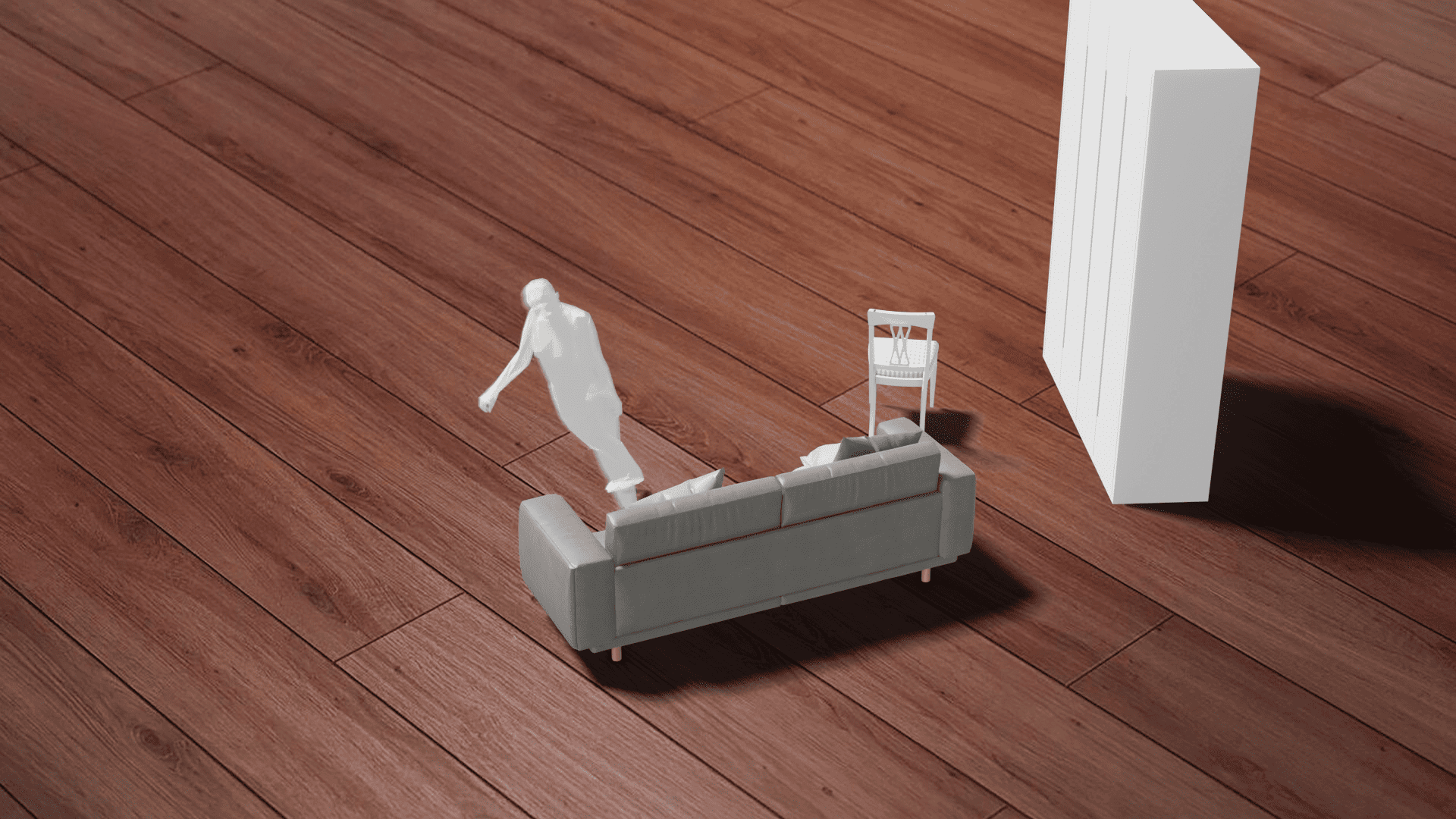} \\
\tiny Miss Contacts (Table)
\end{tabular}
\end{tabular}

\captionof{figure}{
Visualization of a good case scene synthesis compared with failure cases.
}
\label{fig:failure}

\end{minipage}

\vspace{-0.5cm}
\end{figure*}

Figure~\ref{fig:NetworkSimilarity} presents the similarity between features of POSA baseline~\cite{hassan2021populatingPOSA} and features of our ECO model when we keep $1$, $3$, $5$, and all $10$ sparse masks during the inference. We train the ECO model with $K=10$ sparse masks, each mask has a sparsity ratio of $90\%$ in this experiment. The mask score $\bm{\alpha}$ is used to rank and choose useful masks during inference. To compare feature similarity maps, we pass test samples of PROXD dataset~\cite{hassan2019resolvingPROX} to both POSA and our ECO model with the corresponding number of masks. Then, we extract the features from each layer and use the Euclidean distance to compute similarity. While features extracted from the POSA Network remain unchanged in all setups, features of our ECO change when the number of sparse masks is changed.
We can see that in Figure~\ref{fig:NetworkSimilarity}a, using only $1$ mask with the highest mask score $\bm \alpha$ only maintains feature similarity at abstract layers and the dissimilarity significantly increases in later layers (lightens in early layers and darkens in latter ones). Using $3$ masks (Figure~\ref{fig:NetworkSimilarity}b) or $5$ masks (Figure~\ref{fig:NetworkSimilarity}c) shows good feature similarity within corresponding masks (most features show high similarity in their corresponding layers). 
This behavior shows that the representations extracted from each layer in our model are distinctive, highlighting how our proposed method handles redundant information compared with all features from the setup that does not use the mask score $\bm{\alpha}$ to select the useful masks. (Figure~\ref{fig:NetworkSimilarity}d).

\subsection{Failure Cases Analysis }
\label{sec:FailureCases}
Figure~\ref{fig:failure} presents representative failure cases of scene synthesis. Most failures involve objects generated with suboptimal positions, orientations, or incomplete scene layouts, which may reduce the overall plausibility of the synthesized environments. These cases typically arise in highly ambiguous interaction scenarios where multiple scene configurations can satisfy similar contact patterns. Since our method selects a fixed number of sparse masks $\kappa$ during inference, certain interaction cues may receive less emphasis in particularly complex scenes. In addition, some interactions can be supported by multiple plausible object arrangements, making it difficult to recover the exact scene configuration from contact information alone. As a result, the generated scenes may contain redundant objects or local geometric inconsistencies despite preserving the overall interaction structure. Nevertheless, even in these challenging cases, the synthesized scenes generally maintain the major human-scene contact relationships and remain consistent with the intended human activity. These observations suggest that ECO successfully captures interaction-critical information.

\section{Discussion}
\noindent \textbf{Limitations.}
While our method achieves strong performance across multiple human-scene interaction tasks, several limitations remain. First, training with multiple sparse contact masks introduces additional computational overhead compared with dense baselines. Second, the effectiveness of our framework depends on the choice of sparsity ratio and the number of active masks, which may vary across datasets and interaction scenarios. Finally, our current formulation uses a fixed number of selected masks $\kappa$ during inference, which may not fully capture the varying complexity of different interactions.

\noindent \textbf{Future Work.}
Several promising directions remain for future exploration. Learning adaptive mask selection and sparsity ratios directly from data could further improve both efficiency and representation quality. Extending contact-aware sparse representations to temporal interaction modeling may benefit related tasks such as action recognition~\cite{simonyan2014two,wang2016temporal}, pose estimation~\cite{bridgeman2019multi,pavllo20193d}, and object manipulation~\cite{fang2023anygrasp,zhou2022toch}. In addition, integrating contact-aware sparsification with emerging generative frameworks may further improve the scalability of human-scene interaction systems.

\noindent \textbf{Conclusion.}
We present ECO, a contact-aware sparse representation framework for efficient human-scene interaction modeling. Our key insight is that physical interactions are inherently sparse, as only a small subset of human vertices actively participates in contact with the surrounding environment. By explicitly identifying informative contact regions through sparse contact masks and combining them with sparse operators, ECO effectively reduces representation redundancy while preserving interaction-critical information. Extensive experiments on three benchmark datasets demonstrate that our approach consistently improves reconstruction accuracy while substantially reducing inference cost compared with existing methods. These results suggest that contact-aware sparsity provides an effective and general representation paradigm for efficient human-scene interaction reasoning.


%
%
\bibliographystyle{splncs04}
\bibliography{eccvw2026}
\end{document}